\documentclass[10pt,a4paper]{article}
\input{glyphtounicode}
\usepackage[T1]{fontenc}
\usepackage[utf8]{inputenc}
\usepackage[lining]{PlayfairDisplay}
\usepackage[semibold,lining,proportional]{sourceserifpro}
\usepackage[semibold,lining,tabular]{sourcesanspro}
\usepackage[scale=0.86]{sourcecodepro}

\usepackage[activate={true,nocompatibility},final,tracking=false,factor=1100,stretch=12,shrink=12]{microtype}
\usepackage[a4paper,textwidth=136mm,hmarginratio=1:1,top=26mm,bottom=30mm,footskip=14mm]{geometry}
\usepackage{amsmath}
\usepackage{xcolor}
\definecolor{ink}{HTML}{13235A}
\definecolor{cobalt}{HTML}{2548BC}
\definecolor{coral}{HTML}{C8452F}
\definecolor{muted}{HTML}{5C6585}
\definecolor{hair}{HTML}{D5D8E3}
\definecolor{paper}{HTML}{F4F1EA}
\definecolor{wash}{HTML}{EEF1FB}

\usepackage{graphicx}
\usepackage{booktabs,array,colortbl,tabularx}
\usepackage{enumitem}
\usepackage{titlesec}
\usepackage{fancyhdr}
\usepackage[hang,flushmargin,bottom]{footmisc}
\usepackage[labelfont={bf,color=cobalt},textfont={color=ink},font={footnotesize,sf},labelsep=quad,justification=raggedright,singlelinecheck=false,skip=6pt]{caption}
\usepackage{tikz}
\usetikzlibrary{positioning,arrows.meta,calc,fit}
\usepackage{pgfplots}
\pgfplotsset{compat=1.18}
\usepgfplotslibrary{groupplots}
\usepackage[hyphens]{url}
\makeatletter\def\UrlBigBreaks{\do@url@hyp}\makeatother
\usepackage[round,sort,authoryear]{natbib}
\usepackage{hyperref}
\hypersetup{colorlinks=true,linkcolor=cobalt,citecolor=cobalt,urlcolor=cobalt,pdfborder={0 0 0},
  pdfdisplaydoctitle=true,bookmarksnumbered=true}
\setlist{leftmargin=1.35em,itemsep=0.3em,topsep=0.25em,parsep=0pt,partopsep=0pt}
\setlist[itemize]{label={\color{cobalt}\textbullet}}
\setlist[enumerate]{label={\color{cobalt}\sffamily\bfseries\arabic*}}
\titleformat{\section}{\sffamily\fontsize{15}{19}\selectfont\bfseries\color{ink}}{\llap{\textcolor{cobalt}{\thesection}\hspace{3.2mm}}}{0pt}{\phantomsection}
\titleformat{\subsection}{\sffamily\fontsize{11}{14}\selectfont\bfseries\color{ink}}{\llap{\textcolor{muted}{\mdseries\small\thesubsection}\hspace{3.2mm}}}{0pt}{\phantomsection}
\titleformat{\paragraph}[runin]{\sffamily\normalsize\bfseries\color{ink}\spaceskip=0.24em plus 0.08em minus 0.03em\relax}{}{0pt}{}[\enspace]
\titlespacing*{\section}{0pt}{1.9em plus 0.3em minus 0.2em}{0.5em}
\titlespacing*{\subsection}{0pt}{1.25em plus 0.2em minus 0.1em}{0.25em}
\titlespacing*{\paragraph}{0pt}{0.7em}{0pt}

\fancypagestyle{plain}{\fancyhf{}\fancyfoot[R]{\sffamily\scriptsize\color{muted}\thepage}}

\newlength{\widewidth}
\newsavebox{\widebox}
\newenvironment{wide}{\begin{lrbox}{\widebox}\begin{minipage}{\widewidth}\setlength{\parskip}{0pt}}{\end{minipage}\end{lrbox}\noindent\makebox[\textwidth][c]{\usebox{\widebox}}}
\makeatletter\newcommand{\keepspace}[1]{\par\penalty-100\begingroup\setlength{\dimen@}{#1}\dimen@ii\pagegoal\advance\dimen@ii-\pagetotal\ifdim\dimen@>\dimen@ii\break\fi\endgroup}\makeatother

\arrayrulecolor{hair}
\newcommand{\tabfont}{\sffamily\footnotesize\color{ink}}

\newcommand{\toprl}{\arrayrulecolor{ink}\toprule\arrayrulecolor{hair}}
\newcommand{\botrl}{\arrayrulecolor{ink}\bottomrule\arrayrulecolor{hair}}

\newcolumntype{L}{>{\raggedright\arraybackslash}X}
\newcolumntype{P}[1]{>{\raggedright\arraybackslash}p{#1}}

\newlanguage\nohyphenlang
\ExplSyntaxOn
\tl_new:N \l__wtn_tl
\DeclareRobustCommand{\wtn}[1]{{\language\nohyphenlang\ttfamily\mdseries\upshape
  \tl_set:Nn \l__wtn_tl {#1}\tl_replace_all:Nnn \l__wtn_tl {'}{\textquotesingle}\l__wtn_tl}}
\ExplSyntaxOff
\DeclareRobustCommand{\spk}[1]{{\language\nohyphenlang\rmfamily\itshape\mdseries #1}}

\newcommand{\chg}[1]{{\setlength{\fboxsep}{1.2pt}\kern-1.2pt\colorbox{wash}{\strut\textcolor{cobalt}{#1}}\kern-1.2pt}}

\newlength{\barlen}

\providecommand{\doi}{}
\renewcommand*{\doi}{\begingroup\catcode`\_=12\relax\doiaux}
\newcommand*{\doiaux}[1]{doi:\,\href{https://doi.org/#1}{\nolinkurl{#1}}\endgroup}

\pgfplotsset{
  paperaxis/.style={font=\sffamily\footnotesize,axis line style={hair},tick style={hair},
    label style={font=\sffamily\footnotesize,text=ink},tick label style={font=\sffamily\footnotesize,text=ink},
    /pgf/number format/assume math mode=true,grid style={hair!60},
    legend style={font=\sffamily\footnotesize,draw=none,fill=none,text=ink,/tikz/every even column/.append style={column sep=9pt}},
    /pgf/number format/1000 sep={,},scaled ticks=false,tick align=outside,xtick pos=bottom,ytick pos=left,axis lines*=left},
}

\hypersetup{
  pdftitle={Alania-2: Turkish text-to-speech with instant and fine-tuned voice cloning},
  pdfauthor={Sezgin Saygili, Emre Kaplaner, Oncel Ozgul, Fikri San Koktas},
  pdfsubject={Text-to-speech},
  pdfkeywords={text-to-speech, Turkish, voice cloning, synthetic training data, text normalization, low-latency inference},
  pdflang={en}
}

\begin{document}
\thispagestyle{plain}
\begingroup
\setlength{\parskip}{0pt}
{\playfairLF\fontsize{33}{37}\selectfont\color{ink}Alania-2\par}
\vspace{2.5mm}
{\sffamily\fontsize{13}{17}\selectfont\color{ink}Turkish text-to-speech\\ with instant and fine-tuned voice cloning\par}
\vspace{5.5mm}
{\sffamily\small\bfseries\color{ink}Sezgin Saygili\hspace{1.05em}Emre Kaplaner\hspace{1.05em}Oncel Ozgul\hspace{1.05em}Fikri San Koktas\par}
\vspace{1mm}
{\sffamily\footnotesize\color{muted}PatientDesk AI\hspace{1em}\href{mailto:sezgin@patientdesk.ai}{\color{muted}sezgin@patientdesk.ai}\hspace{1em}\href{https://speech.patientdesk.ai}{\color{muted}speech.patientdesk.ai}\par}
\vspace{4.5mm}
{\color{hair}\rule{\textwidth}{0.4pt}\par}
\vspace{2.5mm}
{\sffamily\footnotesize\bfseries\color{cobalt}Abstract\par}
\vspace{1mm}
{\small We present Alania-2, a Turkish text-to-speech (TTS) system with two built-in voices, voice design from text descriptions, and two kinds of voice cloning. Alania-2 is a 488M-parameter model with the VoxCPM architecture, which generates continuous audio latents with a diffusion head on top of a language-model backbone. We train it from scratch on 18,195 hours of Turkish speech, 90\% of which is synthetic speech produced by a larger teacher model and filtered automatically. A text normalizer rewrites numbers, abbreviations and foreign names into the words a Turkish speaker would say; without it, none of 96 test sentences with digits kept all its critical parts, such as an amount or a date, through synthesis and speech recognition, against 86 with it. On Freya-TR-Eval, a public benchmark of 495 Turkish sentences scored with automatic speech recognition, Alania-2 reaches a word error rate of 1.44\%, compared with 5.99\% for our previous system. On its built-in voices, it comes within half a point of character error rate of a 2.3B-parameter model at under a quarter of the size, and a 7.6M-parameter single-voice model trained with the same recipe matches the most intelligible systems we measured, which suggests that training data matters more for intelligibility than model size. We also report practical findings on voice cloning: conditioning on the reference audio alone, rather than continuing it, cuts skipped words from 21.9\% to 1.7\%; 20--30 seconds of reference audio works best; and a low-rank adapter trained in about two minutes raises speaker similarity well above instant cloning. Finally, we describe inference-engine changes that bring first audio down to 32\,ms for a single stream. We release our evaluation harness and three Turkish speech and text datasets.
\par}
\vspace{1.8mm}
{\sffamily\scriptsize\color{muted}\textbf{Keywords:} text-to-speech; Turkish; voice cloning; synthetic training data; text normalization; low-latency inference\par}
\vspace{2.5mm}
{\color{hair}\rule{\textwidth}{0.4pt}\par}
\endgroup

\section{Introduction}\label{sec:introduction}

Turkish is spoken by more than eighty million people, but it remains a mid-resource language for speech synthesis. Recent zero-shot TTS models \citep{chen2025valle,du2024cosyvoice,chen2024f5tts,zhou2025voxcpm} were trained mainly on English and Chinese speech. Multilingual successors such as VoxCPM2 \citep{zhou2026voxcpm2} support Turkish as one of 30 languages, and open Turkish systems differ by an order of magnitude in intelligibility (Section~\ref{sec:evaluation}).

Turkish text is also hard to read aloud correctly. Numbers, dates and foreign names take case suffixes that must agree with how they are pronounced, and the dotted and dotless \emph{i} distinguish words but are easily broken by case conversion designed for English.

We built Alania-2 for telephone voice assistants and long-form narration. These uses need reliable reading of numbers, doses, dates and names; low latency when streaming; consistent built-in voices; and voice cloning that customers can use on their own. Our previous system, Alania-1, is a 3.3B-parameter language model over 24\,kHz SNAC codec tokens \citep{siuzdak2024snac}. It misreads about one word in seventeen on the Freya-TR-Eval benchmark \citep{pamuk2026freyatts}.

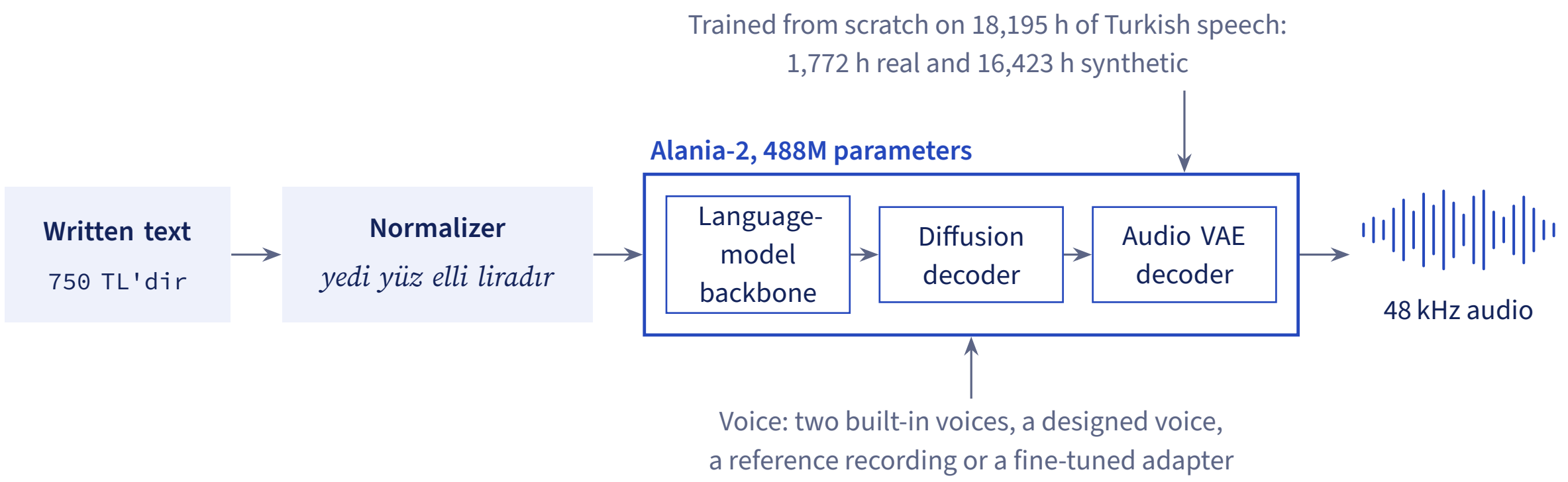
\begin{figure}[t]
\begin{wide}
\centering
\begin{tikzpicture}[font=\sffamily\footnotesize,text=ink,
  blk/.style={fill=wash,inner sep=5pt,minimum height=1.35cm,align=center},
  sub/.style={draw=cobalt,line width=0.5pt,fill=white,inner sep=3pt,minimum height=0.95cm,text width=1.62cm,align=center},
  note/.style={text=muted,align=center},
  arr/.style={-{Stealth[length=1.9mm]},line width=0.6pt,muted}]
\node[blk,text width=1.9cm] (text) {\textbf{Written text}\\[3pt]\wtn{750 TL'dir}};
\node[blk,right=0.5cm of text,text width=2.75cm] (norm) {\textbf{Normalizer}\\[3pt]\spk{yedi yüz elli liradır}};
\node[sub,right=0.72cm of norm] (lm) {Language-model backbone};
\node[sub,right=0.28cm of lm] (dit) {Diffusion decoder};
\node[sub,right=0.28cm of dit] (vae) {Audio VAE decoder};
\node[draw=cobalt,line width=0.9pt,fit=(lm)(dit)(vae),inner sep=6pt,label={[font=\sffamily\footnotesize\bfseries,text=cobalt,anchor=south west,inner sep=2pt]north west:Alania-2, 488M parameters}] (model) {};
\node[right=0.5cm of model,align=center] (audio) {\begin{tikzpicture}[x=0.5mm,y=0.3mm]
  \foreach \x/\h in {0/2,2/4,4/3,6/7,8/10,10/6,12/12,14/8,16/13,18/9,20/5,22/11,24/13,26/6,28/4,30/8,32/11,34/7,36/3,38/2}
    {\draw[cobalt,line width=0.75pt,line cap=round] (\x,-\h) -- (\x,\h);}
  \end{tikzpicture}\\[3pt]48\,kHz audio};
\draw[arr] (text) -- (norm);
\draw[arr] (norm) -- (model.west |- norm);
\draw[arr] (lm) -- (dit);
\draw[arr] (dit) -- (vae);
\draw[arr] (model.east |- audio) -- (audio);
\node[note,below=0.62cm of model] (voice) {Voice: two built-in voices, a designed voice,\\a reference recording or a fine-tuned adapter};
\draw[arr] (voice) -- (model.south);
\node[note,anchor=south east] (train) at ([yshift=0.82cm]model.north east) {Trained from scratch on 18,195 h of Turkish speech:\\1,772 h real and 16,423 h synthetic};
\draw[arr] (train.south -| vae) -- (model.north -| vae);
\end{tikzpicture}
\caption{Overview of Alania-2. Text is first rewritten into the words a speaker would say (Section~\ref{sec:normalization}); the model then generates 48\,kHz audio in the requested voice (Sections~\ref{sec:model} and~\ref{sec:cloning}).}\label{fig:overview}
\end{wide}
\end{figure}

Figure~\ref{fig:overview} gives an overview of Alania-2. Our contributions are:
\begin{enumerate}
  \item \textbf{A data recipe for a mid-resource language.} We train a 488M-parameter model from scratch on 90\% synthetic speech and show that, on its built-in voices, it comes within half a point of character error rate of a fine-tuned model more than four times its size (Sections~\ref{sec:data} and~\ref{sec:model}).
  \item \textbf{A text normalizer} with 134 readings and a lexicon reviewed by a native-speaker author, and respelling of foreign brand names (Section~\ref{sec:normalization}).
  \item \textbf{A comparison with 14 other systems} under one protocol (Section~\ref{sec:evaluation}).
  \item \textbf{Evidence that model size matters little for intelligibility.} A 2.3B-parameter model trained on our data is barely more intelligible than Alania-2, and a 7.6M-parameter single-voice model trained with the same recipe matches the best systems we measured (Section~\ref{sec:size}).
  \item \textbf{Practical findings on voice cloning}: why continuing a reference skips words, how much reference audio helps, and fast fine-tuning with low-rank adapters (Section~\ref{sec:cloning}).
  \item \textbf{Inference-engine changes} for low latency and for serving many fine-tuned voices from one model (Section~\ref{sec:engine}).
\end{enumerate}
We release the Freya-TR-Eval harness and results, a 2,398-hour Turkish caption set, a 3,411-hour synthetic speech set and a 1.44M-sentence domain text corpus. We do not release the model weights.

\section{Related Work}\label{sec:related}

\paragraph{Zero-shot TTS and voice cloning.}
Codec language models clone a voice by continuing a short audio prompt \citep{chen2025valle}. CosyVoice \citep{du2024cosyvoice} combines supervised semantic tokens with flow matching, and F5-TTS \citep{chen2024f5tts} removes the need for explicit text--speech alignment. VoxCPM \citep{zhou2025voxcpm} predicts continuous latents of an audio variational autoencoder (VAE) with a diffusion decoder on a language-model backbone, without a discrete speech tokenizer. VoxCPM2 \citep{zhou2026voxcpm2} extends it to 30 languages, voice design from text and a VAE that encodes at 16\,kHz and decodes at 48\,kHz. We use the VoxCPM2 architecture, tokenizer and VAE but train our own weights. Our previous system, Alania-1, followed the codec route with SNAC tokens \citep{siuzdak2024snac}.

\paragraph{Turkish TTS.}
\citet{yeshpanov2023multilingual} train a Tacotron~2 model on Kazakh alone and synthesize nine other Turkic languages, including Turkish, by mapping their letters through the International Phonetic Alphabet to Kazakh letters. FreyaTTS \citep{pamuk2026freyatts} is a 183M-parameter Turkish-first flow-matching model released together with the Freya-TR-Eval benchmark, and \citet{ozkose2026comprehensive} compare four recent systems on Turkish, fine-tuned and zero-shot, with eighteen objective metrics.

\paragraph{Data and evaluation tools.}
We build our real-speech set from YODAS \citep{li2024yodas}, a multilingual corpus of YouTube audio, much of it with manual or automatic subtitles. We filter it with a language classifier trained on VoxLingua107 \citep{valk2020voxlingua107}, the AST audio tagger \citep{gong2021ast}, DNSMOS quality prediction \citep{reddy2021dnsmos} and Whisper speech recognition \citep{radford2022whisper}. Our synthetic text comes from FineWeb2 \citep{penedo2025fineweb2}, Wikipedia and sentences written by Qwen3 \citep{yang2025qwen3}. For evaluation, we measure intelligibility with Whisper, naturalness with UTMOS \citep{saeki2022utmos}, and speaker similarity with ECAPA-TDNN embeddings \citep{desplanques2020ecapa,ravanelli2021speechbrain}; FLEURS \citep{conneau2022fleurs} supplies some of the cloning references.

\paragraph{Efficient serving.}
PagedAttention \citep{kwon2023vllm} manages the key/value cache of language models in fixed-size blocks, and S-LoRA \citep{sheng2023slora} and Punica \citep{chen2023punica} serve many low-rank adapters (LoRA; \citealp{hu2021lora}) on one base model. Our engine builds on Nano-vLLM \citep{nanovllm,nanovllmvoxcpm}, which uses the same block-based cache.

\section{Data}\label{sec:data}

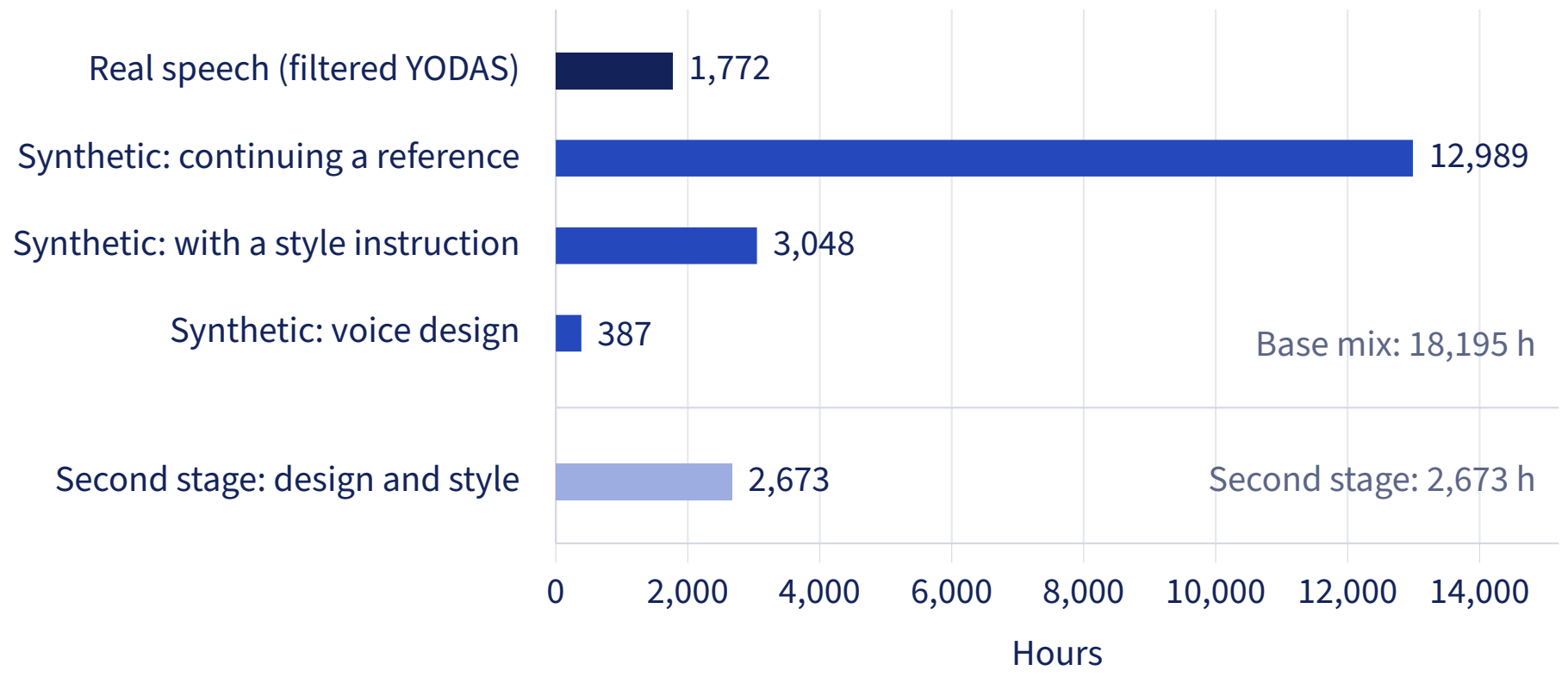
\begin{figure}[t]
\centering
\begin{tikzpicture}
\begin{axis}[paperaxis,xbar,scale only axis,width=7.7cm,height=4.1cm,bar width=8pt,xmin=0,xmax=15200,xlabel={Hours},y dir=reverse,
  ytick={0,1,2,3,4.7},yticklabels={Real speech (filtered YODAS),Synthetic: continuing a reference,Synthetic: with a style instruction,Synthetic: voice design,Second stage: design and style},
  ymin=-0.7,ymax=5.4,ytick style={draw=none},xmajorgrids,
  nodes near coords,nodes near coords style={font=\sffamily\footnotesize,text=ink,anchor=west}]
\addplot[fill=ink,draw=none,bar shift=0pt] coordinates {(1772,0)};
\addplot[fill=cobalt,draw=none,bar shift=0pt] coordinates {(12989,1) (3048,2) (387,3)};
\addplot[fill=cobalt!45,draw=none,bar shift=0pt] coordinates {(2673,4.7)};
\draw[hair] (axis cs:0,3.85) -- (axis cs:15200,3.85);
\node[font=\sffamily\footnotesize,text=muted,anchor=east] at (axis cs:15100,3.15) {Base mix: 18,195 h};
\node[font=\sffamily\footnotesize,text=muted,anchor=east] at (axis cs:15100,4.7) {Second stage: 2,673 h};
\end{axis}
\end{tikzpicture}
\caption{Training data. The base mix is 90\% synthetic speech; a short second stage adds voice design and style instructions.}\label{fig:data}
\end{figure}

Figure~\ref{fig:data} summarizes the training data. The base mix combines 1,772 hours of filtered real speech with 16,423 hours of synthetic speech; a second, shorter stage adds voice design and style instructions.

\subsection{Real speech}
We start from the Turkish part of YODAS \citep{li2024yodas}: 38,494 recordings totalling 9,216 hours. Most of this audio is weakly labelled. Of the recordings, 97\% carry automatic YouTube captions, and only 2,448 hours are labelled as Turkish at all. We therefore treat every file as untrusted. We split it into segments, check the language with a classifier trained on VoxLingua107 \citep{valk2020voxlingua107} and with Whisper \citep{radford2022whisper}, remove music and singing with the AST audio tagger \citep{gong2021ast}, reject clipped or band-limited audio, and keep segments with a DNSMOS score of at least 2.4 \citep{reddy2021dnsmos}. Finally, we transcribe every segment again with Duyu, our Turkish speech recognizer, a fine-tune of Whisper large-v3 turbo.

This leaves 2,398 hours (835,267 segments, median length 9.9\,s). Qwen3-32B \citep{yang2025qwen3} writes a short style caption for each segment, describing the speaker and the delivery. For training, we apply stricter filters and remove segments with a low DNSMOS score (189,631), a Duyu transcript that disagrees with the YouTube caption (42,186), or an implausible speaking rate (152). The training set holds 598,213 segments (1,772 hours) from 19,969 speaker clusters; we hold out about 1\% of the clusters for validation.

\subsection{Synthetic speech}
Turkish speech with the quality, speaker balance and text coverage we needed is not available at this scale, so we generate it with VoxCPM2 \citep{zhou2026voxcpm2}, using 10 diffusion steps and a guidance scale of 2.0. One H100 GPU produces 83 seconds of audio per second.

\paragraph{Text.} The text comes from FineWeb2 \citep{penedo2025fineweb2} (40\%), domain sentences written by Qwen3-30B-A3B \citep{yang2025qwen3} (25\%), Wikipedia (22.5\%) and YODAS transcripts (12.5\%). After removing duplicates, 5.65M of 6.18M text units remain. Each unit is normalized (Section~\ref{sec:normalization}) with the normalizer version current at the time.

\paragraph{Voices.} The speakers are our two built-in voices and a bank of 2,752 designed voices (1,283 female, 1,469 male). We create the bank from 5,000 text descriptions and drop near-duplicates whose speaker embeddings have a cosine similarity of 0.85 or more. Each clip is generated in one of three modes: continuing a reference recording (77\%), following a reference together with an English style instruction (20\%), or designing a voice from its description alone (3\%). About 22.5\% of the training clips are also passed through a simulated room, microphone, noise and codec channel.

\paragraph{Quality control.} We keep a clip only if it passes every check in Table~\ref{tab:qc}. In a manual audit of 681 generated clips, 79.2\% passed.

\begin{table}[t]
\centering
\tabfont
\caption{Checks applied to every synthetic clip. A clip is kept only if it passes all of them.}\label{tab:qc}
\begin{tabular*}{\linewidth}{@{\extracolsep{\fill}}lll@{}}
\toprl
Check & Measured with & Threshold \\
\midrule
Intelligibility & CER, faster-whisper large-v3 & $\leq$ 0.05 \\
Speaking rate & characters per second & 7--25 \\
Voice match & ECAPA similarity to the reference & $\geq$ 0.45 ($\geq$ 0.35 with a style instruction) \\
Naturalness & UTMOS22 & $\geq$ 2.5 \\
Pitch & median F0 & consistent with the voice's gender \\
\botrl
\end{tabular*}
\end{table}

\subsection{Second stage and released data}
The second-stage mix has 1.4M rows: voice descriptions, short paraphrased descriptions, style-instruction rows, and 400,000 plain rows that keep the model's base behaviour. We release three datasets on Hugging Face:
\begin{itemize}
  \item \textbf{Captions}\footnote{\url{https://huggingface.co/datasets/cloud0day3/alania-speech-captions-tr}}: transcripts and style captions for the 835,267 real segments, without audio.
  \item \textbf{Synthetic speech}\footnote{\url{https://huggingface.co/datasets/cloud0day3/alania-synthetic-speech-tr}}: 3,411 hours at 48\,kHz, designed voices only.
  \item \textbf{Domain sentences}\footnote{\url{https://huggingface.co/datasets/cloud0day3/alania-domain-text-tr}}: 1.44M sentences.
\end{itemize}

\section{Text Normalization}\label{sec:normalization}

Alania-2 reads Turkish spelling and has no phoneme input, so numbers, symbols, abbreviations and foreign words must first be rewritten into the words a speaker would say. Table~\ref{tab:tnexamples} shows typical cases. Our normalizer is a sequence of rules for numbers, money, dates, times, ranges, phone numbers, doses, units, titles, addresses and symbols. After expanding a number, an amount or a unit, it re-attaches the Turkish suffix that followed it and adjusts the suffix to the new word, mainly its vowels (vowel harmony). A companion paper describes the normalizer in detail and evaluates its version of 7 October 2026, which has 131 test cases; here we summarize the parts that matter most for the model.

\begin{table}[t]
\centering
\tabfont
\caption{Examples of written input and the text Alania-2 receives.}\label{tab:tnexamples}
\begin{tabular*}{\linewidth}{@{\extracolsep{\fill}}lll@{}}
\toprl
Case & Written & Spoken form given to the model \\
\midrule
Money with a suffix & \wtn{750 TL'dir} & \spk{yedi yüz elli liradır} \\
Decimal amount & \wtn{12,50 TL} & \spk{on iki lira elli kuruş} \\
Percentage with a suffix & \wtn{\%2,5'lik} & \spk{yüzde iki buçukluk} \\
Turkish initialism & \wtn{SGK} & \spk{segeka} \\
Borrowed initialism & \wtn{CEO} & \spk{siio} \\
Brand name & \wtn{PatientDesk} & \spk{Peyşınt Desk} \\
Brand name & \wtn{ChatGPT} & \spk{Çet Ci Pi Ti} \\
\botrl
\end{tabular*}
\end{table}

\paragraph{Initialisms and native review.} Turkish initialisms are usually read as Turkish letter names joined into one word. Some are read as words, and a few borrowed ones use English letter names. A native-speaker author reviewed 134 written forms in 13 categories, initialisms among them, on a page that showed each planned reading as text. The author accepted 118 as shown and changed 16: 10 marked wrong and 6 marked right with a better reading typed in.

\paragraph{Upper case.} Words written entirely in capitals are lowercased with Turkish rules, and every remaining \emph{I} then becomes \emph{ı} and every \emph{İ} becomes \emph{i}. Other capitals stay. A model that keeps capital letters may otherwise read a capital \emph{I} the English way: in an earlier test of ours, VoxCPM2 read the city name \emph{Iğdır} as \emph{iğdir}.

\paragraph{Foreign words and brands.} Read with Turkish spelling rules, some brand names and English words come out garbled. A word seen at least 1,000 times in the text behind our 225,764-word Turkish word list stays as written. Otherwise we respell it in Turkish letters if it has an inner capital, or if it is an English word, meaning ASCII letters only, present in the CMU Pronouncing Dictionary\footnote{\url{https://github.com/cmusphinx/cmudict}} and absent from the word list. The respelling comes from the word's pronunciation in that dictionary, and these automatic outputs have not had a native review.

\paragraph{Lexicon.} Automatic respelling is not always better than the written form, so we checked 100 candidate brand and English terms. We generated each term both as written and respelled, in one carrier sentence, and transcribed both with Duyu. A native-speaker author listened to both versions of every term and decided 94 of them, giving a lexicon of 17 respelled and 77 unchanged terms; the other 6 were left to the automatic rule. Users can add their own entries. Table~\ref{tab:respell} gives the scores: for the 17 respelled terms, recognition similarity rises from 0.63 to 0.99. We used this study to build the lexicon, not to test it. Each form of a term has one render, and a rule using these same scores proposed 16 of the 17 respellings, so the rise is partly built in.

\begin{table}[t]
\centering
\tabfont
\caption{Recognition of 100 brand and English terms by Duyu, in one carrier sentence with one voice and seed. Similarity is between the part of the transcript that fills the term's slot and either the term or its respelling, whichever is higher (0--1) $\uparrow$.}\label{tab:respell}
\begin{tabular*}{\linewidth}{@{\extracolsep{\fill}}lcc@{}}
\toprl
 & As written & Respelled \\
\midrule
Mean similarity, all 100 terms & 0.911 & \textbf{0.961} \\
Transcript equals the term or its respelling & 72 & \textbf{85} \\
Mean similarity, the 17 respelled terms & 0.627 & \textbf{0.989} \\
\botrl
\end{tabular*}
\end{table}

\paragraph{Effect of normalizing at all.} Normalization is not optional for Alania-2. The companion paper tests this on 150 constructed sentences in the style of clinic and customer-service messages, covering numbers, dates, units, phone numbers, initialisms, brands and addresses. Most of their expected spoken forms have not had a native review. We ran the model directly, with one voice and one render per sentence, and transcribed the audio with Duyu. Given raw written text, the model had a word error rate of 66.7\%, against 6.4\% with normalized text. A critical span is a part of a sentence that must survive, such as an amount or a name. Every critical span was recovered from the transcript in 7\% of sentences with raw text and in 81\% with normalized text. Among the 96 sentences containing digits, this held for none with raw text and for 86 with normalized text. In the transcripts, numbers come out as other words. Nobody listened to these renders, so this description rests on transcripts and audio length.

This describes Alania-2, which was trained on normalized text, not neural TTS in general. Zero-shot VoxCPM2, which generated most of Alania-2's training audio, did not break down on raw text in a separate evaluation. Over 24 prompts and two seeds, its character error rate (CER) was 10.9\% on raw text against 5.6\% on normalized text on prompts built to stress normalization, and 7.7\% against 5.5\% on numeric prompts. That evaluation used other prompts, another recognizer and another scorer, so these figures cannot be set against the ones above number by number, but they suggest that the failure on digits is not inherent to the architecture.

\paragraph{Effect of respelling.} Respelling adds a smaller gain whose size is uncertain, on terms already used to build the lexicon. Among the sentences whose text it changes, every critical span was recovered in 12 of 15 with respelling and 3 of 15 without it through our speech API, and in 11 and 6 of 16 when we ran the model directly.

\section{Model}\label{sec:model}

\subsection{Choosing an architecture}
\begin{figure}[t]
\centering
\begin{tikzpicture}
\begin{axis}[paperaxis,scale only axis,width=9.4cm,height=4.8cm,xmode=log,xmin=0.0045,xmax=0.6,ymin=2.9,ymax=4.08,
  xlabel={CER (mean, log scale), lower is better},ylabel={UTMOS22, higher is better},xmajorgrids,ymajorgrids,clip=false,
  log ticks with fixed point,xtick={0.01,0.1},xticklabels={0.01,0.1}]
\addplot[only marks,mark=*,mark size=2.2pt,muted!55] coordinates {(0.358,3.021) (0.156,3.061) (0.075,3.716) (0.018,3.843)};
\addplot[only marks,mark=*,mark size=2.5pt,cobalt] coordinates {(0.0162,3.953) (0.0161,3.897)};
\node[font=\sffamily\footnotesize,text=ink,anchor=south] at (axis cs:0.358,3.04) {Alania-1};
\node[font=\sffamily\footnotesize,text=ink,anchor=east] at (axis cs:0.145,3.061) {Alania-1, revised normalizer};
\node[font=\sffamily\footnotesize,text=ink,anchor=west] at (axis cs:0.08,3.716) {MOSS-TTS};
\node[font=\sffamily\footnotesize,text=ink,anchor=west] at (axis cs:0.02,3.835) {MOSS-TTS fine-tuned};
\node[font=\sffamily\footnotesize,text=cobalt,anchor=west] at (axis cs:0.0176,3.965) {VoxCPM2};
\node[font=\sffamily\footnotesize,text=cobalt,anchor=east] at (axis cs:0.0150,3.897) {VoxCPM2 + LoRA};
\end{axis}
\end{tikzpicture}
\caption{Choosing an architecture: intelligibility against naturalness on 120 Turkish prompts, 240 renders per system. The best systems sit at the top left.}\label{fig:base}
\end{figure}
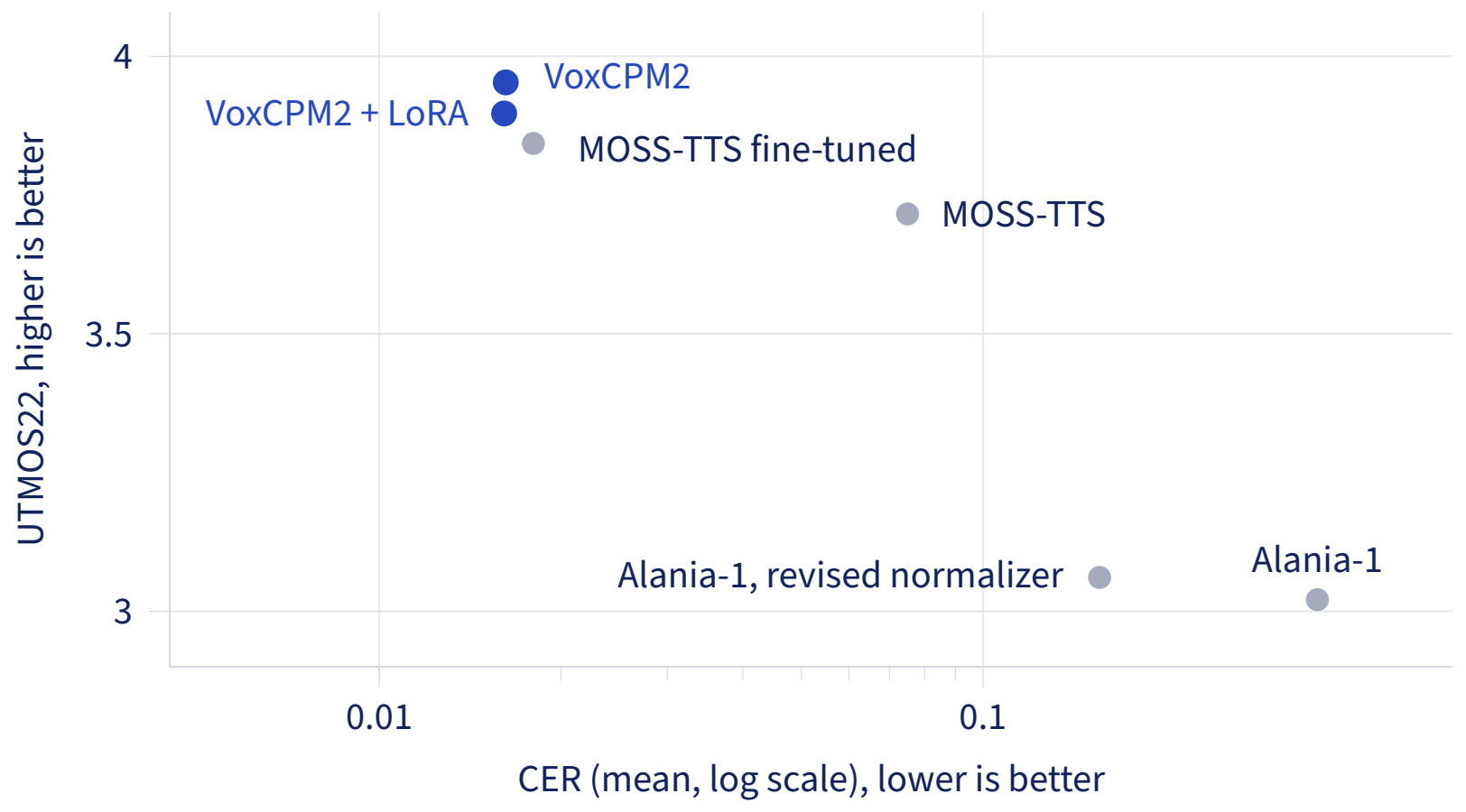

We compared candidate architectures on 120 Turkish prompts with two seeds each (Figure~\ref{fig:base}). Zero-shot VoxCPM2 was the most intelligible (CER 0.016), the most natural by UTMOS (3.95) and the most consistent between renders of the same voice (speaker similarity 0.83). Unlike Alania-1, which ran on past the end of the text in 7.5\% of renders, it never did. Its VAE also preserved speaker identity through encoding and decoding far better than the SNAC codec (similarity 0.93 against 0.71). We therefore adopted the VoxCPM architecture.

\subsection{Architecture and training}
Alania-2 keeps VoxCPM2's text tokenizer and its frozen audio VAE \citep{zhou2026voxcpm2}, which encodes 16\,kHz audio into 64-dimensional latents at 25 frames per second and decodes them to 48\,kHz. A language-model backbone reads the text and the latents generated so far. A diffusion transformer then predicts the next patch of four latent frames (0.16\,s of audio) in 10 flow-matching steps with classifier-free guidance. Our model is smaller than VoxCPM2 (Table~\ref{tab:arch}), and we train all its other weights from random initialization with the VoxCPM training code.

\begin{table}[t]
\centering
\tabfont
\caption{Alania-2 architecture, excluding the frozen audio VAE. Attention heads stay 128-dimensional to match VoxCPM's position encoding.}\label{tab:arch}
\begin{tabular*}{\linewidth}{@{\extracolsep{\fill}}lccr@{}}
\toprl
Component & Layers $\times$ width & Heads & Params \\
\midrule
Backbone LM & 16 $\times$ 1024 & 8 (2 KV) & 268M \\
Residual LM & 6 $\times$ 1024 & 8 (2 KV) & 72M \\
Feature encoder & 8 $\times$ 768 & 6 & 69M \\
Diffusion decoder & 8 $\times$ 768 & 6 & 72M \\
\midrule
Total & & & 488M \\
\botrl
\end{tabular*}
\end{table}

We pursued two tracks in parallel. \textbf{Track 1} continued training the full VoxCPM2 on our data (2.3B parameters in the released checkpoint, excluding the VAE). \textbf{Track 2} trained our own smaller model from scratch. A model trained from scratch needs many steps before it learns to align text with audio: a 1.24B-parameter variant never did within 54k steps (CER 0.84--0.88), while the 488M model did. The released model was trained in three stages:
\begin{enumerate}
  \item about 64k steps on an earlier 8,904-hour mix;
  \item 70k steps on the 18,195-hour base mix, with a batch of 32 per GPU, two gradient-accumulation steps, a learning rate of $10^{-4}$ and weight decay of 0.05;
  \item the second stage on design and style data at a learning rate of $5\times10^{-5}$, stopped at step 4,000.
\end{enumerate}
Training took on the order of 210 H200 GPU-hours, not counting data generation.

\begin{figure}[t]
\centering
\begin{tikzpicture}
\begin{axis}[paperaxis,ybar,area legend,scale only axis,width=12.2cm,height=4.2cm,bar width=7.5pt,ymin=0,ymax=15,ylabel={CER (\%)},
  symbolic x coords={V2,T1a,T1b,T2a,T2b,A2},xtick=data,
  xticklabels={{VoxCPM2\\2.3B},{Track 1, round 1\\2.3B},{Track 1, round 2\\2.3B},{Track 2, first run\\488M},{Track 2, base\\488M},{\textbf{Alania-2}\\488M}},
  xticklabel style={align=center},xtick style={draw=none},enlarge x limits=0.09,ymajorgrids,
  legend style={at={(0,1.04)},anchor=south west,legend columns=3}]
\addplot[fill=cobalt,draw=none] coordinates {(V2,2.7) (T1a,2.8) (T1b,2.4) (T2a,6.1) (T2b,2.7) (A2,2.7)};
\addplot[fill=cobalt!45,draw=none] coordinates {(V2,2.9) (T1a,2.7) (T1b,2.3) (T2a,5.5) (T2b,2.8) (A2,2.8)};
\addplot[fill=coral,draw=none] coordinates {(V2,2.2) (T1a,2.2) (T1b,2.0) (T2a,14.2) (T2b,9.0) (A2,3.4)};
\legend{Female voice,Male voice,With a style instruction}
\end{axis}
\end{tikzpicture}
\caption{CER of the candidate models on our 292-clip suite (one seed; Whisper large-v3), lower is better. On its built-in voices, Alania-2 is close to the 2.3B models at under a quarter of their size, and the second training stage fixed most of the base model's errors with style instructions. Full results are in Appendix~\ref{app:tracks}.}\label{fig:tracks}
\end{figure}
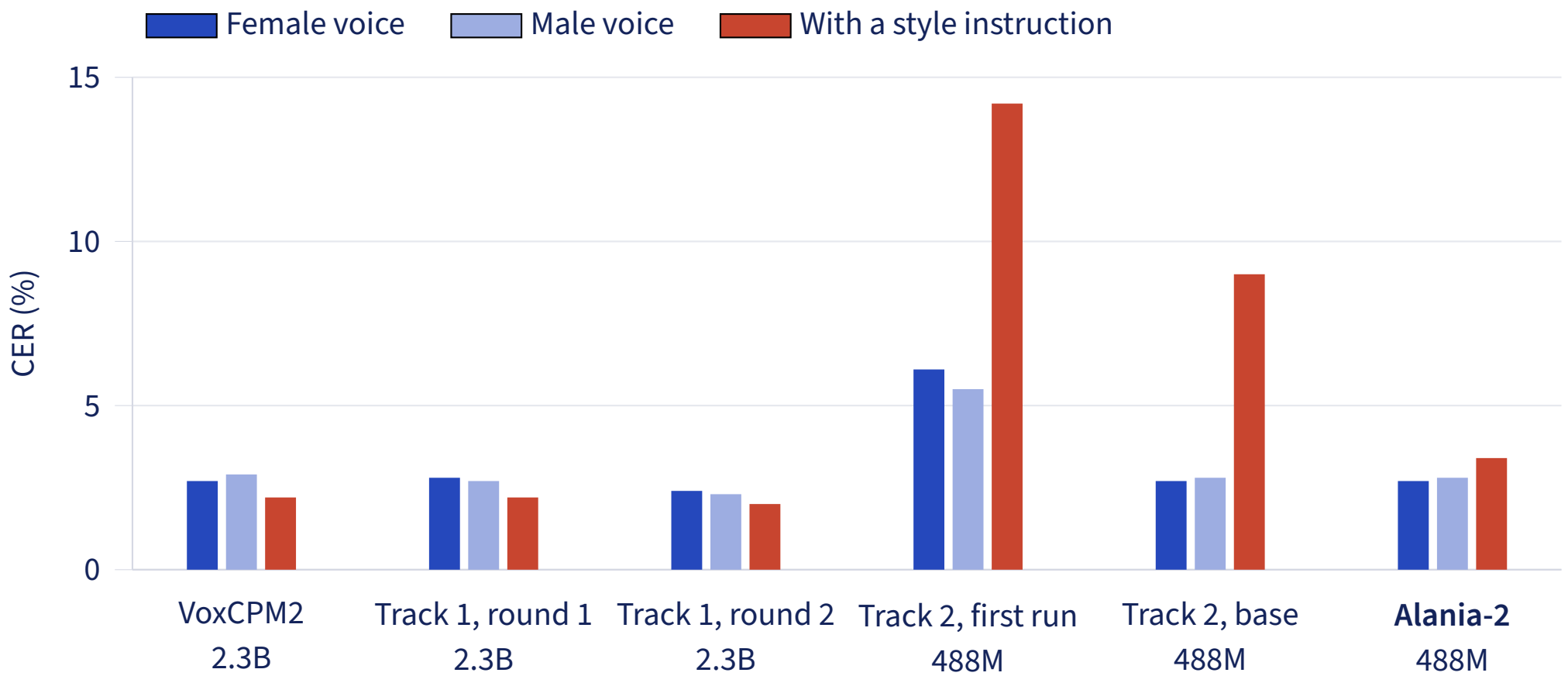

\subsection{Comparison and choice}
Figure~\ref{fig:tracks} compares the candidates. On the two built-in voices, Alania-2's CER is within half a point of VoxCPM2 and of both Track 1 models. With style instructions its CER is higher (3.4\% against 2.0--2.2\% for Track 1). It produced the requested gender for 30 of 32 voice designs, as often as any candidate. Its UTMOS is 0.2--0.3 lower than VoxCPM2's, and it is weaker at cloning a voice from an English recording (Section~\ref{sec:cloning}). We chose Alania-2 because it is far cheaper to serve at nearly the same intelligibility: on one H200, it keeps up with real time at 12 requests per second, against about 8 for Track 1.

\subsection{Voices, styles and design}
The two built-in voices are designed voices, not recordings of real people. We chose them in a blind comparison of 12 candidates. For each voice we store a reference clip, an average speaker embedding and a pitch range (199.6--238.6\,Hz for the female voice, 109.1--131.6\,Hz for the male voice). At serving time, we use the pitch range to reject renders that drift away from the voice.

Style control comes from the captioned training rows. At first, the model sometimes read the instruction aloud: Turkish instructions on every take, and long English ones on 4 of 6. We now rewrite each instruction into a short English tag with a small LLM, after which all 54 test takes were clean.

Ready-made styles (warm, calm, energetic, smiling) keep CER between 0.33\% and 1.82\% and widen the pitch range by 0.5--1.4 semitones. Whispering and slow speech are beyond the model. For voice design we use a guidance scale of 3.0, chosen from a sweep between 1.5 and 4.0 in which higher guidance raised UTMOS (from 1.79 to 2.93) but lowered gender accuracy (from 30 to 27 of 32).

\section{Evaluation}\label{sec:evaluation}

\subsection{Freya-TR-Eval}
Freya-TR-Eval \citep{pamuk2026freyatts} is a public Turkish benchmark of 495 sentences of 3--13 words, given as raw written text. It measures intelligibility: each system reads every sentence, a speech recognizer transcribes the audio, and the benchmark reports the word error rate (WER) against the original text. Its released file contains no digits, symbols, initialisms or mixed-case brand names, so it does not exercise text normalization. We re-ran it for 15 systems under one protocol, plus the bare Alania-2 model without our serving checks, and released the harness and results. Appendix~\ref{app:systems} lists the exact checkpoint of every system.\footnote{\url{https://huggingface.co/datasets/PatientdeskAI/freya-tr-eval-rerun}} Each system reads every sentence three times. We band-limit the audio to 8\,kHz, resample it to 16\,kHz, transcribe it with Whisper large-v3 \citep{radford2022whisper} through faster-whisper\footnote{\url{https://github.com/SYSTRAN/faster-whisper}} (beam size 5) and score it with the benchmark's own text normalization. We also report UTMOS22 on the full-band audio and the real-time factor on one L4 GPU.

\begin{figure}[t]
\centering
\begin{tikzpicture}
\begin{axis}[paperaxis,xbar,scale only axis,width=8.3cm,height=7.3cm,bar width=6.5pt,xmin=0,xmax=13.5,
  xlabel={Word error rate (\%), lower is better},ytick={0,...,15},y dir=reverse,ytick style={draw=none},
  yticklabels={EMA Lightning (8.6M),Antalia-2 Mini (7.6M),Trendyol-TTS (2.38B),Gemini 3.8 Flash TTS,Gemini 3.8 Flash-Lite TTS,Anka TTS (336M),\textbf{Alania-2, API (488M)},\textbf{Alania-2, model only},Chatterbox Multilingual (500M),Piper (16M),XTTS-v2 (470M),Alania-1 (3.3B),Antalia-1 (305M),MMS-TTS (36M),Coqui GlowTTS (28M),FreyaTTS-small (183M)},
  enlarge y limits=0.035,xmajorgrids,
  nodes near coords,nodes near coords style={font=\sffamily\footnotesize,text=ink,anchor=west},
  every node near coord/.append style={/pgf/number format/fixed,/pgf/number format/fixed zerofill,/pgf/number format/precision=2}]
\addplot[fill=muted!45,draw=none,bar shift=0pt] coordinates {(1.04,0) (1.16,2) (1.23,3) (1.34,4) (1.42,5) (1.91,8) (3.41,9) (3.67,10) (10.93,14) (12.02,15) (6.38,13)};
\addplot[fill=cobalt!45,draw=none,bar shift=0pt] coordinates {(1.05,1) (5.99,11) (6.28,12)};
\addplot[fill=cobalt,draw=none,bar shift=0pt] coordinates {(1.44,6) (1.51,7)};
\end{axis}
\end{tikzpicture}
\caption{Freya-TR-Eval word error rate (495 sentences, three readings each, faster-whisper large-v3). Our models are in blue. Alania-2 through our API is seventh of the 15 systems; the bare model without serving checks is shown for reference. Antalia-2 Mini is our 7.6M-parameter single-voice model (Section~\ref{sec:size}).}\label{fig:freya}
\end{figure}
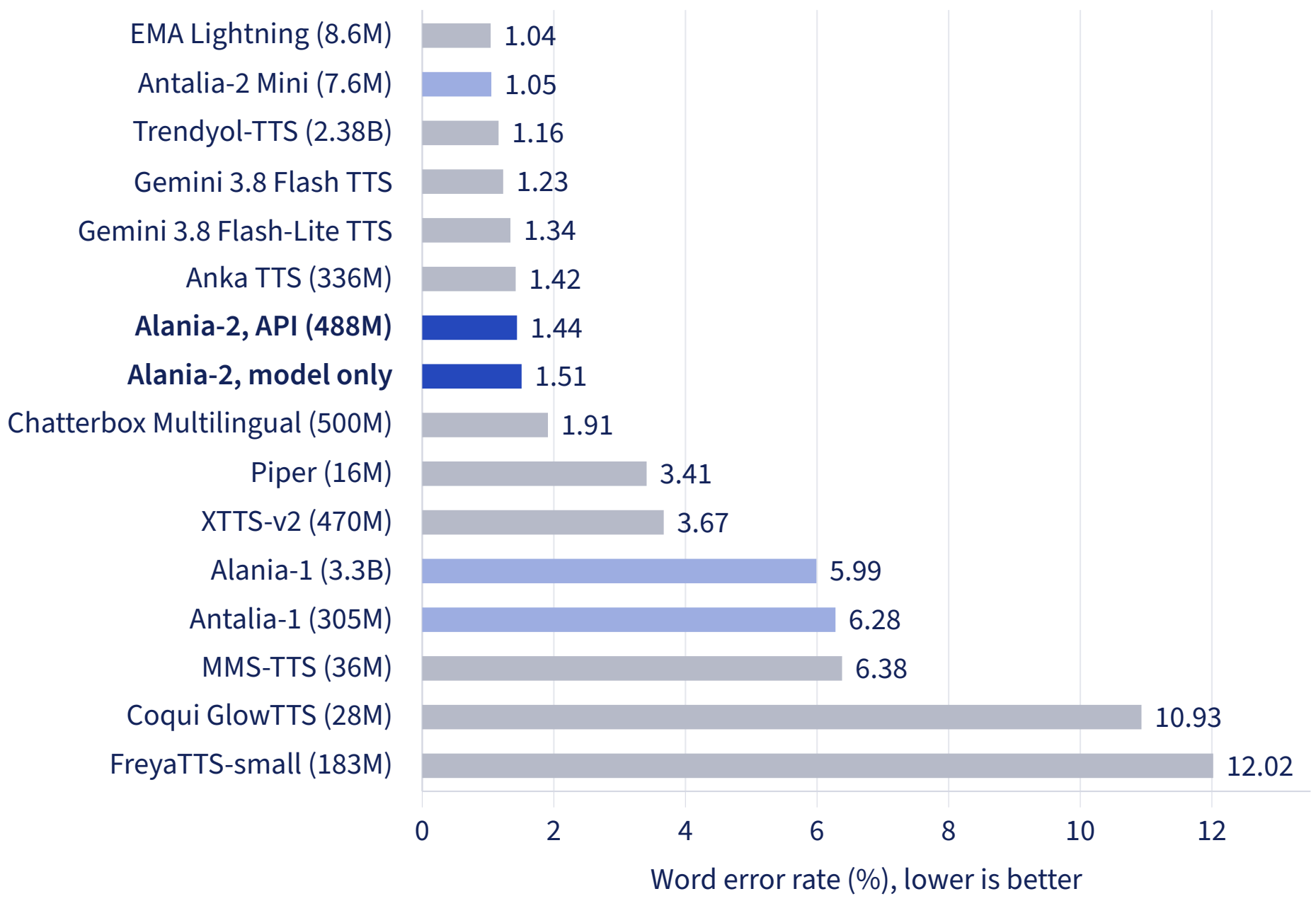

Figure~\ref{fig:freya} shows the results. Through our API, Alania-2 reaches 1.44\% WER, seventh of the 15 systems. The two best are compact models of under 9M parameters: EMA Lightning at 1.04\% and our Antalia-2 Mini at 1.05\%. The benchmark authors report 1.43\% for ElevenLabs v4, which we did not run. Differences of a tenth of a point are within the spread between the three readings. Our serving checks and retries (Section~\ref{sec:serving}) lower the bare model's 1.51\% to 1.44\%. Compared with Alania-1, Alania-2 cuts WER four-fold and character error rate (CER) six-fold (Table~\ref{tab:v1v2}).

\begin{table}[t]
\centering
\tabfont
\caption{Alania-1 and Alania-2 on Freya-TR-Eval, both through our API. RTF is the real-time factor on one L4 GPU (Alania-1 measured on 100 sentences).}\label{tab:v1v2}
\begin{tabular*}{\linewidth}{@{\extracolsep{\fill}}lcc@{}}
\toprl
 & Alania-1 & Alania-2 \\
\midrule
Architecture & LLM over SNAC tokens & VoxCPM \\
Parameters & 3.3B & 488M \\
Sample rate & 24 kHz & 48 kHz \\
WER (\%) $\downarrow$ & 5.99 & \textbf{1.44} \\
CER (\%) $\downarrow$ & 1.81 & \textbf{0.28} \\
UTMOS22 $\uparrow$ & 3.19 & \textbf{3.67} \\
RTF $\downarrow$ & 2.49 & \textbf{0.151} \\
\botrl
\end{tabular*}
\end{table}

\paragraph{System version.} This report describes release \texttt{alania-v2} of October 2026, with two built-in voices and 10 diffusion steps. The service has changed since then. Sampling moved to 6 diffusion steps, and on 11 October release \texttt{alania-v2.1}, a fine-tune of \texttt{alania-v2} for control of speaking style with the same architecture, replaced the model. On 12 October nine voices were added. The Freya-TR-Eval results here use the female voice (\texttt{alania2-f1}) for Alania-2 and version 1.0.0 of Antalia-2 Mini; our released results also hold later runs of both.

\subsection{What we did not measure}
This report does not yet include a listening test with many listeners. Decisions that needed listening, such as choosing the architecture and the built-in voices, were made in blind comparisons by one native speaker, and our only naturalness measure is UTMOS22, which was trained on English. Intelligibility results depend on the recognizer and on how text is normalized before scoring, so we name the recognizer with every result.

\section{How Large Does the Model Need to Be?}\label{sec:size}

Alania-2 has 488M parameters, while open speech models of several billion parameters are available, so it is fair to ask why we did not build on a larger one. We did try: in our experiments, the larger models were barely more intelligible, and a far smaller model trained with the same recipe matched the best systems we measured.

\subsection{Larger models trained on our data}
Table~\ref{tab:size} compares Alania-2 with the larger models from Section~\ref{sec:model}. Continuing to train the full VoxCPM2, at 2.3B parameters, improved CER on the built-in voices by only 0.3--0.5 points over Alania-2. The larger model did keep clear advantages elsewhere. It followed style instructions more accurately (CER 2.0\% against 3.4\%), it sounded somewhat more natural (UTMOS 0.16 higher on the female voice), and it cloned voices from English recordings far better (similarity 0.64 against 0.41, measured before Alania-2's second training stage). Training a larger model from scratch did not help either: our 1.24B-parameter variant had still not learned to align text with audio after 54k steps, when the 488M model already had. At serving time, the smaller model handles 1.5 times as many requests per GPU.

\begin{table}[t]
\begin{wide}
\centering
\tabfont
\caption{Model size against quality on our 292-clip suite (one seed; CER from Whisper large-v3, lower is better) and serving capacity. Training: ``continued'' means further training of VoxCPM2 on our data, ``scratch'' random initialization. Requests/s is the request rate at which about 99\% of streams keep ahead of real time on one H200, higher is better.}\label{tab:size}
\begin{tabular*}{\linewidth}{@{\extracolsep{\fill}}lrcccccc@{}}
\toprl
 & & & \multicolumn{3}{c}{CER (\%) $\downarrow$} & & \\
\cmidrule(lr){4-6}
Model & Params & Training & Female & Male & Style & UTMOS $\uparrow$ & Requests/s $\uparrow$ \\
\midrule
VoxCPM2, zero-shot & 2.3B & released & 2.7 & 2.9 & 2.2 & 3.77 & -- \\
Track 1, round 2 & 2.3B & continued & \textbf{2.4} & \textbf{2.3} & \textbf{2.0} & 3.73 & 8 \\
Track 2-L & 1.24B & scratch & \multicolumn{5}{c}{did not align within 54k steps (CER 84--88\%)} \\
\textbf{Alania-2} & 488M & scratch & 2.7 & 2.8 & 3.4 & 3.57 & \textbf{12} \\
\botrl
\end{tabular*}
\end{wide}
\end{table}

\subsection{Size and intelligibility across systems}
\begin{figure}[t]
\centering
\begin{tikzpicture}
\begin{axis}[paperaxis,scale only axis,width=11.6cm,height=5.8cm,xmode=log,ymode=log,xmin=4,xmax=7000,ymin=0.7,ymax=16,
  xlabel={Parameters, log scale},ylabel={Freya-TR-Eval WER (\%), log scale},
  xtick={10,100,1000},xticklabels={10M,100M,1B},ytick={1,2,5,10},yticklabels={1,2,5,10},xmajorgrids,ymajorgrids,clip=false]
\addplot[only marks,mark=*,mark size=2.2pt,muted!55] coordinates {(8.6,1.04) (2380,1.16) (336,1.42) (500,1.91) (16,3.41) (470,3.67) (36,6.38) (28,10.93) (183,12.02)};
\addplot[only marks,mark=*,mark size=2.5pt,cobalt!45] coordinates {(7.62,1.05) (3300,5.99) (305,6.28)};
\addplot[only marks,mark=*,mark size=2.9pt,cobalt] coordinates {(488,1.44)};
\node[font=\sffamily\footnotesize,text=ink,anchor=south west] at (axis cs:8.6,1.08) {EMA Lightning};
\node[font=\sffamily\footnotesize,text=cobalt,anchor=north west] at (axis cs:7.0,0.99) {Antalia-2 Mini};
\node[font=\sffamily\footnotesize,text=ink,anchor=north] at (axis cs:2380,1.1) {Trendyol-TTS};
\node[font=\sffamily\footnotesize,text=ink,anchor=east] at (axis cs:305,1.42) {Anka};
\node[font=\sffamily\footnotesize\bfseries,text=cobalt,anchor=north west] at (axis cs:500,1.38) {Alania-2};
\node[font=\sffamily\footnotesize,text=ink,anchor=west] at (axis cs:550,1.91) {Chatterbox};
\node[font=\sffamily\footnotesize,text=ink,anchor=west] at (axis cs:17.5,3.41) {Piper};
\node[font=\sffamily\footnotesize,text=ink,anchor=west] at (axis cs:520,3.67) {XTTS-v2};
\node[font=\sffamily\footnotesize,text=cobalt,anchor=south] at (axis cs:3300,6.4) {Alania-1};
\node[font=\sffamily\footnotesize,text=cobalt,anchor=north] at (axis cs:305,5.9) {Antalia-1};
\node[font=\sffamily\footnotesize,text=ink,anchor=west] at (axis cs:40,6.38) {MMS-TTS};
\node[font=\sffamily\footnotesize,text=ink,anchor=east] at (axis cs:25.5,10.93) {GlowTTS};
\node[font=\sffamily\footnotesize,text=ink,anchor=west] at (axis cs:200,12.02) {FreyaTTS-small};
\end{axis}
\end{tikzpicture}
\caption{Freya-TR-Eval word error rate against model size for the 13 systems with published sizes (both axes logarithmic). Our models are in blue. Size does not predict intelligibility.}\label{fig:sizewer}
\end{figure}
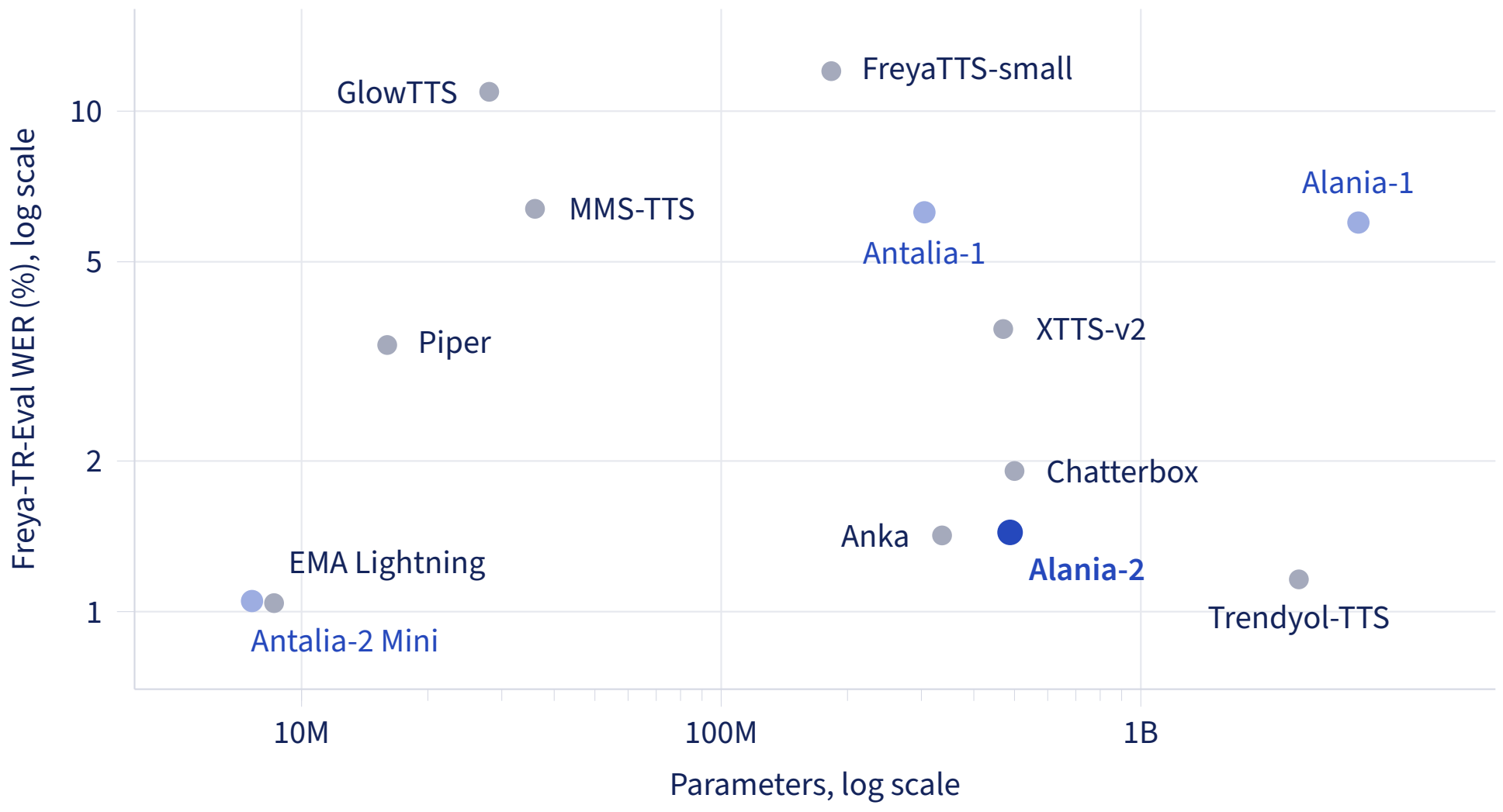

The same holds across the systems in our Freya-TR-Eval re-run (Figure~\ref{fig:sizewer}). Model size does not predict word error rate: the three most intelligible systems with published sizes span 7.6M to 2.4B parameters, and our 3.3B-parameter Alania-1 misreads more than four times as many words as the 488M Alania-2. Training data and other design choices separate these systems far more than size does. This benchmark cannot show the effect of text handling, because its sentences contain no digits, initialisms or mixed-case brand names.

\subsection{Going much smaller}
To test how far the recipe scales down, we trained Antalia-2 Mini, an openly released model with a single voice. It has 7.62M parameters: a 3.69M-parameter acoustic model that generates mel spectrograms with few-step flow matching, and a 3.93M-parameter vocoder that produces 48\,kHz audio. Like Alania-2, it learns from synthetic speech, here about 1,790 hours of one voice generated by VoxCPM2, and it reads text through the same normalizer (Section~\ref{sec:normalization}). On Freya-TR-Eval it reaches 1.05\% WER and 0.20\% CER, the lowest CER of the 15 systems we ran. Its teacher, VoxCPM2 speaking the same voice, scored 0.92\% WER and 0.20\% CER in a single run on our development harness. A model about 300 times smaller thus keeps nearly all of its teacher's intelligibility.

Antalia-2 Mini also shows what the extra parameters pay for. It has one voice and no cloning, voice design or style control, and it sounds less natural than Alania-2 (UTMOS 3.25 against 3.67). For reading plain Turkish sentences in a fixed voice, a few million parameters suffice when the data are right. The larger capacity of Alania-2 goes into flexibility: many voices, cloning from a short recording, voice design and style instructions. We chose 488M as the smallest size we tried that kept these abilities while serving the most requests per GPU.

\section{Voice Cloning}\label{sec:cloning}

Alania-2 offers two kinds of cloned voice. An \textbf{instant clone} is conditioned on a short reference recording at request time. A \textbf{professional clone} is a LoRA adapter trained for one voice on one to five minutes of speech. The reference recordings in this section come from consenting contributors to a Turkish speech corpus and from FLEURS \citep{conneau2022fleurs}. We measure CER and skipped words with Whisper large-v3 and word alignment, and speaker similarity as the cosine similarity of ECAPA embeddings between the generated audio and a held-out recording of the same speaker.

\subsection{Continuing a reference skips words}
VoxCPM can use a reference recording in two ways. In \textbf{continuation} mode, the reference audio and its transcript come before the new text, and the model continues the recording. In \textbf{reference-only} mode, the reference audio sets the voice and no transcript is given.

\begin{figure}[t]
\centering
\begin{tikzpicture}
\begin{axis}[paperaxis,ybar,area legend,scale only axis,width=11.6cm,height=4.2cm,bar width=13pt,ymin=0,ymax=38,ylabel={Percent},
  symbolic x coords={skipall,skipclean,cerclean,cerhard},xtick=data,
  xticklabels={{Skipped words\\all references},{Skipped words\\clean references},{CER\\clean references},{CER\\hard references}},
  xticklabel style={align=center},xtick style={draw=none},enlarge x limits=0.16,ymajorgrids,
  nodes near coords,nodes near coords style={font=\sffamily\footnotesize,text=ink},every node near coord/.append style={/pgf/number format/fixed,/pgf/number format/precision=2},
  legend style={at={(0,1.04)},anchor=south west,legend columns=2}]
\addplot[fill=muted!45,draw=none] coordinates {(skipall,21.9) (skipclean,21.3) (cerclean,25.1) (cerhard,32.6)};
\addplot[fill=cobalt,draw=none] coordinates {(skipall,1.7) (skipclean,0.18) (cerclean,3.1) (cerhard,13.8)};
\legend{Continuation mode (before),Reference-only mode with per-chunk checks (after)}
\end{axis}
\end{tikzpicture}
\caption{Streamed instant clones before and after the fix; lower is better. 200 streams each, with the same seeds and texts of 121--249 characters. ``Hard'' references contain reverberation, music, a laptop microphone, noise or MP3 compression.}\label{fig:clonefix}
\end{figure}
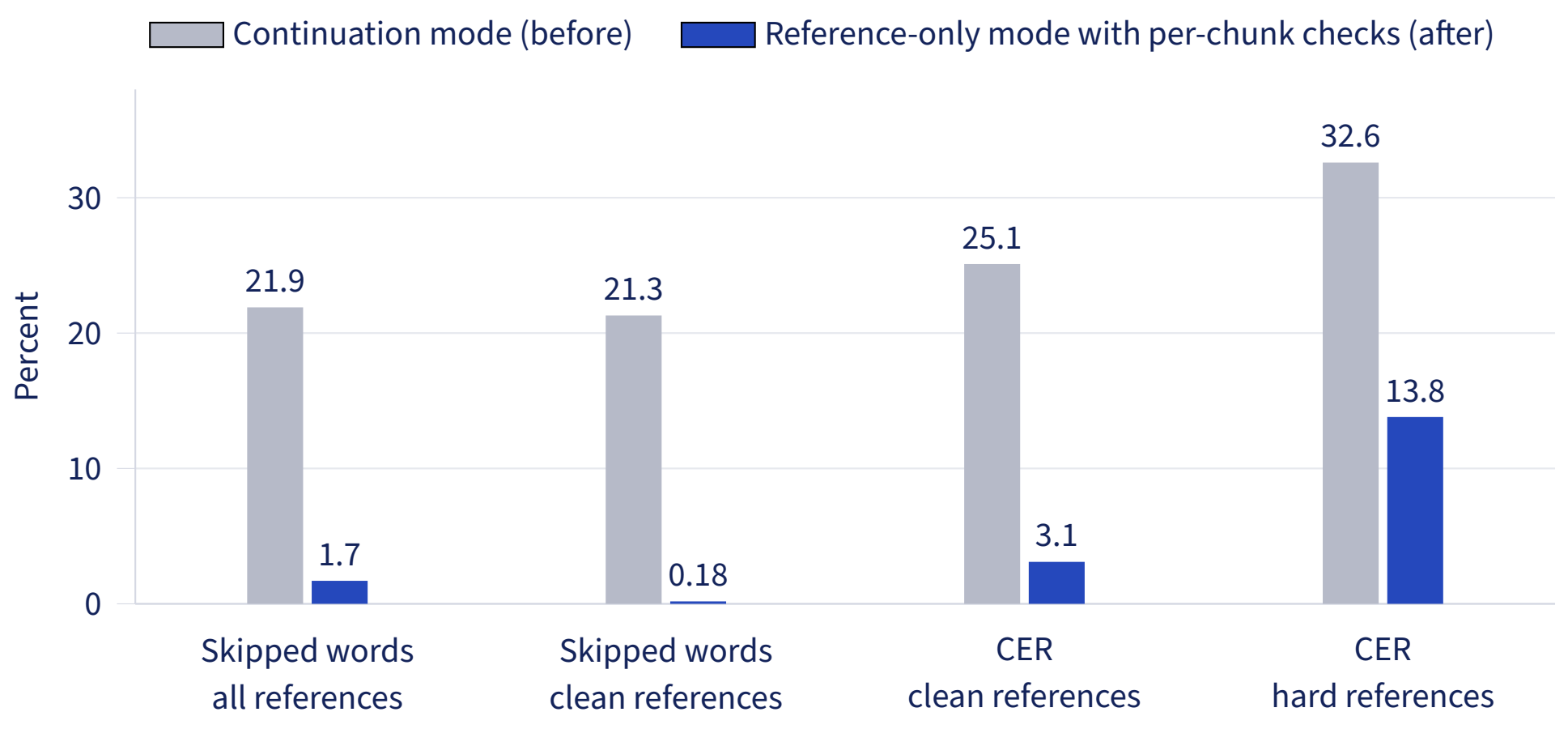

We first served clones in continuation mode, and users reported skipped words, silent gaps and loops. With references of 20\,s or more, continuation skipped about half of every text, often whole chunks, and a wrong reference transcript had the same effect at any length. Switching to reference-only mode, together with per-chunk checks (Section~\ref{sec:serving}), fixed this without losing speaker identity (Figure~\ref{fig:clonefix}). Over 200 streams with the same seeds and texts, skipped words fell from 21.9\% to 1.7\%, and to 0.18\% on clean references. Speaker similarity on clean references was unchanged (0.569 against 0.568). The near-silent chunks that users heard as gaps disappeared (from 45 to 0), and the longest pause where two chunks join mid-sentence fell from 1.04\,s to 0.37\,s.

\subsection{How much reference audio helps}
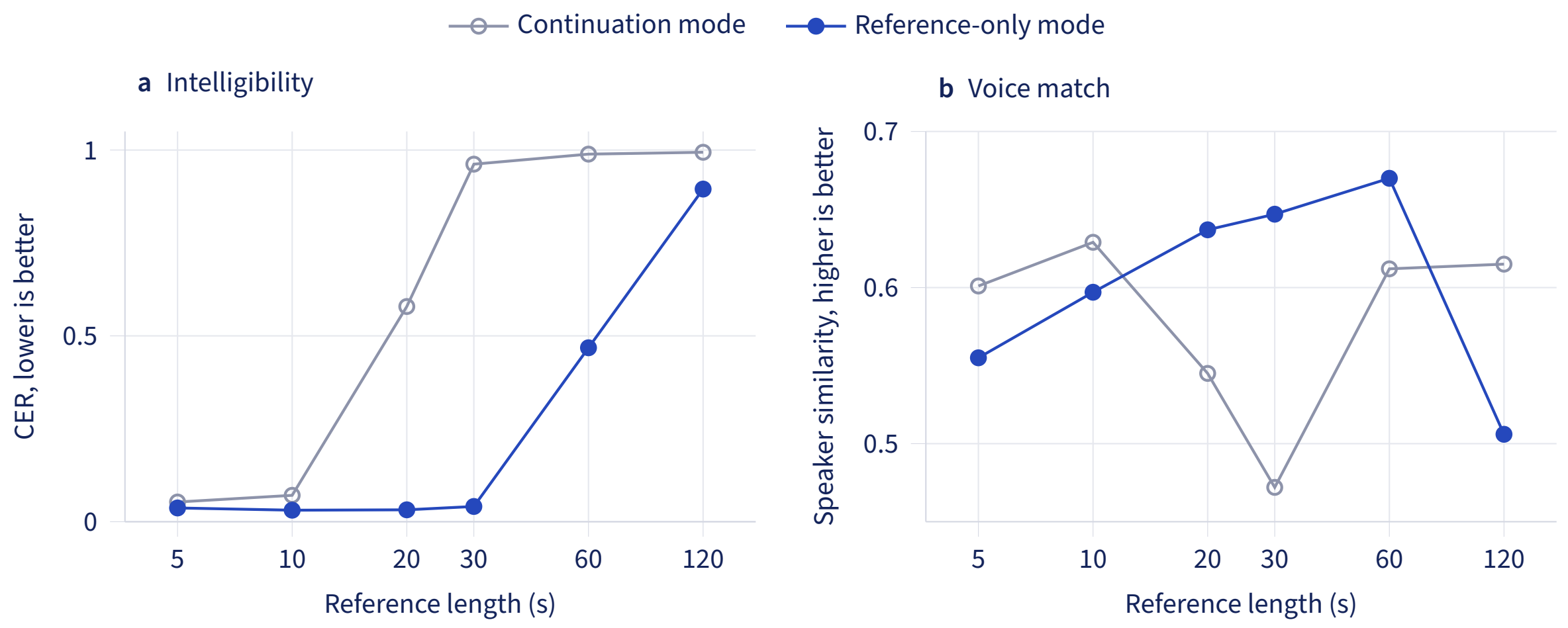
\begin{figure}[t]
\begin{wide}
\centering
\ref{reflenleg}\\[1pt]
\begin{tikzpicture}
\begin{groupplot}[group style={group size=2 by 1,horizontal sep=1.7cm},paperaxis,scale only axis,width=6.3cm,height=3.9cm,
  xmode=log,log basis x=2,xtick={5,10,20,30,60,120},xticklabels={5,10,20,30,60,120},xlabel={Reference length (s)},xmajorgrids,ymajorgrids]
\nextgroupplot[ylabel={CER, lower is better},ymin=0,ymax=1.05,legend to name=reflenleg,legend columns=2,title={\textbf{a}\enspace Intelligibility},title style={at={(0,1)},anchor=south west,font=\sffamily\footnotesize,text=ink}]
\addplot[muted!70,mark=o,thick,mark options={fill=white}] coordinates {(5,0.053) (10,0.071) (20,0.579) (30,0.962) (60,0.989) (120,0.994)};
\addplot[cobalt,mark=*,thick] coordinates {(5,0.037) (10,0.031) (20,0.032) (30,0.041) (60,0.468) (120,0.895)};
\legend{Continuation mode,Reference-only mode}
\nextgroupplot[ylabel={Speaker similarity, higher is better},ymin=0.45,ymax=0.7,title={\textbf{b}\enspace Voice match},title style={at={(0,1)},anchor=south west,font=\sffamily\footnotesize,text=ink},yticklabel style={/pgf/number format/fixed,/pgf/number format/precision=2}]
\addplot[muted!70,mark=o,thick,mark options={fill=white}] coordinates {(5,0.601) (10,0.629) (20,0.545) (30,0.472) (60,0.612) (120,0.615)};
\addplot[cobalt,mark=*,thick] coordinates {(5,0.555) (10,0.597) (20,0.637) (30,0.647) (60,0.670) (120,0.506)};
\end{groupplot}
\end{tikzpicture}
\caption{Instant clones as the reference grows, using the first $N$ seconds of each recording. Reference-only mode is stable up to 30\,s.}\label{fig:reflen}
\end{wide}
\end{figure}

Figure~\ref{fig:reflen} compares the two modes as the reference grows from 5 to 120\,s (four speakers, three texts, three seeds). In reference-only mode, similarity rises from 0.56 at 5\,s to 0.64--0.67 at 20--30\,s. Beyond that, the model breaks down, with a CER of 0.47 at 60\,s and 0.90 at 120\,s. Continuation mode is already unusable at 20\,s. Choosing the cleanest 20-second window of a recording is better than taking its first 20 seconds (similarity 0.672 against 0.637, CER 0.024 against 0.032). We therefore accept references of 3--30\,s, recommend 20--30\,s, and automatically pick the cleanest 20-second window from longer uploads. Encoding the reference takes 31\,ms at 5\,s and 404\,ms at 120\,s, while first audio stays between 83 and 114\,ms.

\paragraph{Chaining chunks does not help.} We also tried conditioning each chunk of a long text on the end of the previous chunk. Over 80 paired renders, this lowered CER by 0.035--0.069 but also lowered similarity by 0.058--0.074, and the voice drifted from chunk to chunk. Every chunk therefore stays conditioned on the original reference.

\paragraph{Reference quality.} We convert saved references to mono 48\,kHz at $-20$\,dBFS and check them for background noise, echo, clipping and implausible transcripts. Background noise only 12\,dB below the speech made the model skip 29--49\% of words.

\subsection{Professional clones}
A professional clone is a rank-32 LoRA adapter \citep{hu2021lora} on the backbone and the diffusion decoder. The training pipeline splits the upload into 3--15\,s utterances with a signal-to-noise ratio above 15\,dB, transcribes them, holds out 10\% for evaluation, and trains for 200 steps on 80\,s of speech, rising linearly to 400 steps for five minutes. The adapter is used only if it beats the instant clone made from the best 20-second reference on the held-out speech: similarity must rise by at least 0.02, and CER must stay below $\max(0.05, 1.5c)$, where $c$ is the instant clone's CER.

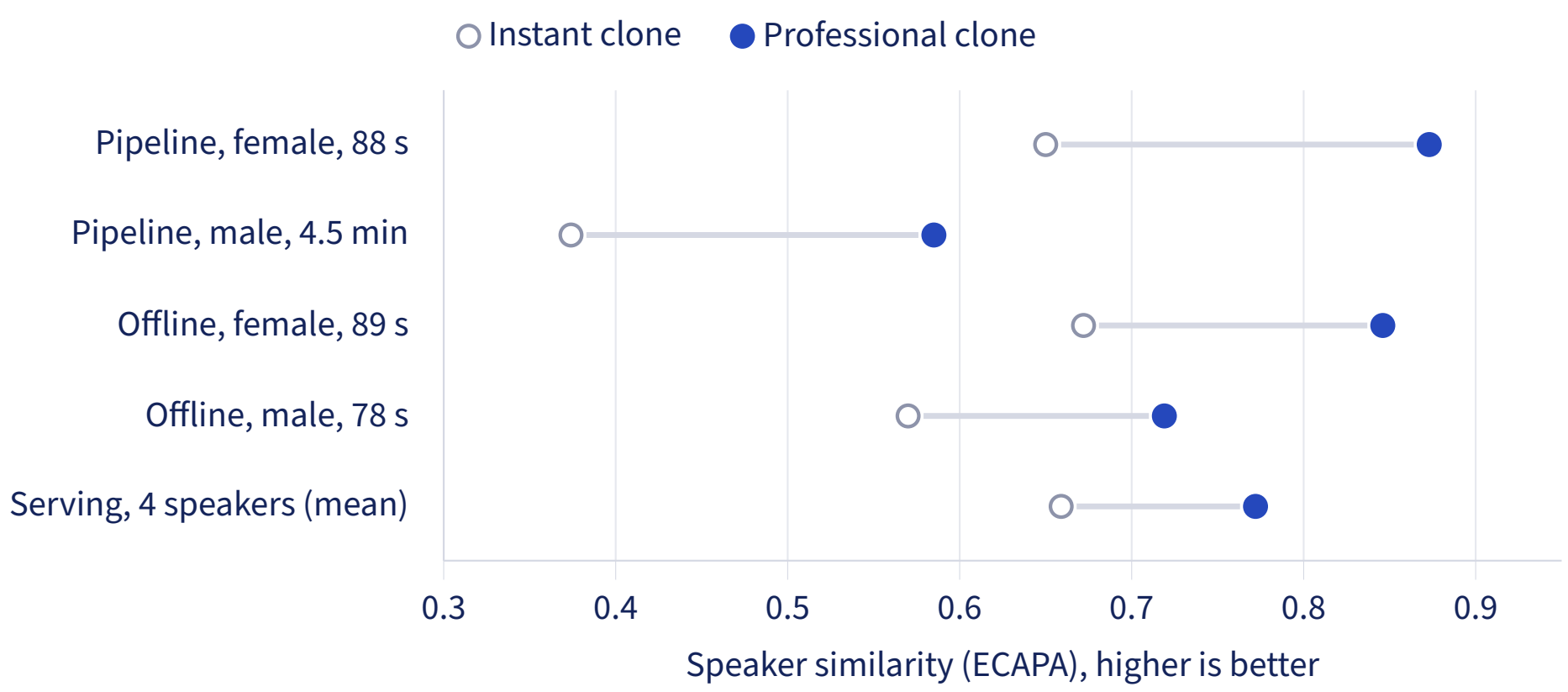
\begin{figure}[t]
\centering
\begin{tikzpicture}
\begin{axis}[paperaxis,scale only axis,width=8.8cm,height=3.7cm,xmin=0.3,xmax=0.95,xlabel={Speaker similarity (ECAPA), higher is better},
  ytick={1,2,3,4,5},yticklabels={{Pipeline, female, 88 s},{Pipeline, male, 4.5 min},{Offline, female, 89 s},{Offline, male, 78 s},{Serving, 4 speakers (mean)}},
  ytick style={draw=none},ymin=0.4,ymax=5.6,y dir=reverse,xmajorgrids,xticklabel style={/pgf/number format/fixed,/pgf/number format/precision=1},
  legend style={at={(0,1.04)},anchor=south west,legend columns=2}]
\addplot[only marks,mark=o,mark size=2.4pt,thick,muted!70,mark options={fill=white}] coordinates {(0.650,1) (0.374,2) (0.672,3) (0.570,4) (0.659,5)};
\addplot[only marks,mark=*,mark size=2.6pt,cobalt] coordinates {(0.873,1) (0.585,2) (0.846,3) (0.719,4) (0.772,5)};
\foreach \a/\b/\y in {0.650/0.873/1,0.374/0.585/2,0.672/0.846/3,0.570/0.719/4,0.659/0.772/5}{\edef\tmp{\noexpand\draw[hair,line width=1.3pt] (axis cs:\a+0.009,\y) -- (axis cs:\b-0.009,\y);}\tmp}
\legend{Instant clone,Professional clone}
\end{axis}
\end{tikzpicture}
\caption{Speaker similarity of professional clones (LoRA adapters) against instant clones of the same speakers, on held-out speech. Each row is one run; the last is the mean over four speakers through the serving engine.}\label{fig:lora}
\end{figure}

Training takes 84--126\,s on one A100, and a whole job finishes in 3--4.5 minutes. Figure~\ref{fig:lora} shows large similarity gains in every run. One speaker's adapter trained on 580\,s of speech (similarity 0.68) scored below the same speaker's 78-second adapter (0.72), which suggests that matching the recording session matters more than the amount of audio at this scale. Cloning a voice from an English recording into Turkish speech remains weak: similarity 0.41, against 0.64 for Track 1, measured before the second training stage.

\subsection{Verification}
Because a professional clone reproduces a voice closely, we require the voice's owner to be present. The speaker reads a random Turkish sentence aloud in the browser. The sentence is drawn from about three million combinations and expires after 15 minutes. Duyu must recognize the sentence, and the speaker embedding of the reading must match the uploaded recordings. On 13 consenting speakers (84 same-speaker pairs across recordings and 2,460 different-speaker pairs), a threshold of 0.45 rejects 0.5\% of genuine readings and accepts 0.45\% of impostor pairs. In a live test, a different speaker's reading scored 0.20 and was rejected within 3\,s. Our API also refuses to synthesize any pending challenge sentence, so the check cannot be passed by generating the reading with an instant clone through our own service.

\section{Serving}\label{sec:serving}

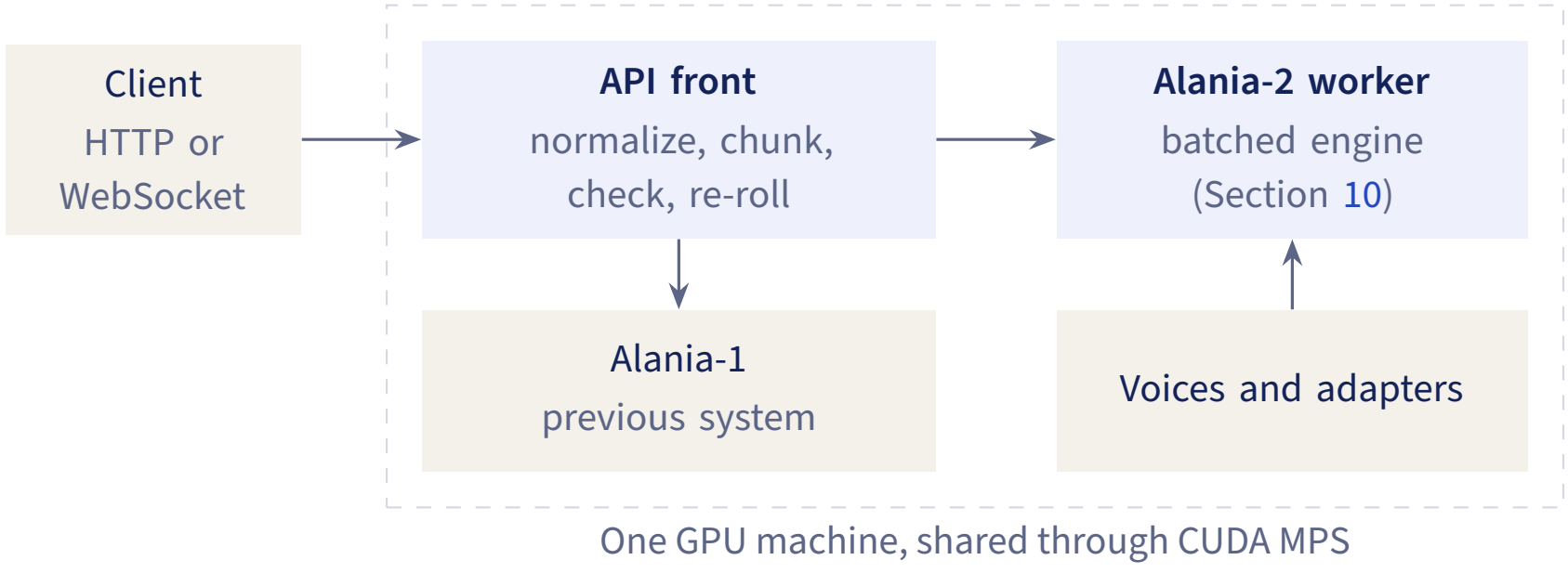
\begin{figure}[t]
\centering
\begin{tikzpicture}[font=\sffamily\footnotesize,text=ink,
  box/.style={fill=wash,inner sep=5pt,minimum height=1.15cm,align=center},
  plain/.style={fill=paper,inner sep=5pt,minimum height=1.15cm,align=center},
  arr/.style={-{Stealth[length=1.9mm]},line width=0.6pt,muted}]
\node[plain,text width=1.75cm] (client) {Client\\[1pt]{\color{muted}HTTP or WebSocket}};
\node[box,right=0.85cm of client,text width=3.3cm] (front) {\textbf{API front}\\[1pt]{\color{muted}normalize, chunk, check, re-roll}};
\node[box,right=0.85cm of front,text width=3.0cm] (worker) {\textbf{Alania-2 worker}\\[1pt]{\color{muted}batched engine (Section~\ref{sec:engine})}};
\node[plain,below=0.5cm of front,text width=3.3cm] (v1) {Alania-1\\[1pt]{\color{muted}previous system}};
\node[plain,below=0.5cm of worker,text width=3.0cm] (store) {Voices and adapters};
\draw[arr] (client) -- (front);
\draw[arr] (front) -- (worker);
\draw[arr] (front) -- (v1);
\draw[arr] (store) -- (worker);
\node[draw=hair,dashed,fit=(front)(worker)(v1)(store),inner sep=7pt,label={[font=\sffamily\footnotesize,text=muted]below:One GPU machine, shared through CUDA MPS}] {};
\end{tikzpicture}
\caption{Serving layout on each GPU machine.}\label{fig:serving}
\end{figure}

Each GPU machine runs the Alania-2 worker and an API front that also serves Alania-1 (Figure~\ref{fig:serving}). CUDA's Multi-Process Service (MPS) lets the two models share the GPU fairly: without it, four Alania-1 streams next to an Alania-2 load fell from 100\% to 3\% keeping up with real time; with it, they held 100\% next to 24 Alania-2 streams.

\paragraph{From text to audio.} The front normalizes the text (Section~\ref{sec:normalization}) and splits it into chunks of at most 140 characters, preferring sentence and then clause boundaries. It then resolves the voice (built-in, designed, instant clone or professional clone) and streams the first chunk as it is generated, in steps of 0.16\,s.

\paragraph{Checks and retries.} Later chunks are generated up to two chunks ahead of playback and checked before they are sent. A chunk is rejected if it is empty; if it runs on too long (more than 2.0\,s plus 0.11\,s per character); if its length does not fit its text; if it contains a pause of more than 1\,s; if it has no voiced speech; or, for the built-in voices, if its median pitch is outside the voice's range. A rejected chunk is generated again with a new seed, up to twice, and the best attempt is kept. If waiting would make playback stall, the chunk is sent unchecked. When joining chunks, we trim their edge silence and insert 350\,ms of silence after a sentence and 120\,ms after a clause. Every response is labelled as AI-generated in its headers.

\paragraph{Deployment.} We run on spot RTX PRO 6000 machines in one EU region, with on-demand A100 machines as a floor that is never preempted. A new machine serves Alania-2 81\,s after boot. Professional clones are trained on a separate, always-on GPU machine in the EU.

\section{Inference Engine}\label{sec:engine}

The worker runs Alania-2 on Nano-vLLM-VoxCPM \citep{nanovllmvoxcpm}, an inference engine for VoxCPM based on Nano-vLLM \citep{nanovllm}. It batches concurrent streams and keeps the backbone's key/value cache in fixed-size blocks \citep{kwon2023vllm}. We changed the engine in three places to cut first-audio latency and raise throughput, and in several more to serve many fine-tuned voices from one model. We checked each change against the unmodified engine on the same seeds and texts.

\subsection{Latency and throughput}

Table~\ref{tab:engine} lists the three changes:
\begin{itemize}
  \item \textbf{Batched VAE decoding.} The audio decoder ran once per stream. Running it once per batch cut decoding at 96 streams from 110 to 27\,ms per step. At 48 streams, throughput rose from 41 to 74 seconds of audio per second, and the share of streams that kept ahead of real time rose from none to all.
  \item \textbf{CUDA-graph prefill.} We capture the prompt step as CUDA graphs, as the engine already did for later steps. Single-stream first audio fell from 74 to 32\,ms, and the output stayed bit-identical (23 of 23 renders).
  \item \textbf{Kernel warm-up.} A batch size the engine had not seen before stalled every stream for about 1.4\,s. We now run the common batch sizes once at start-up.
\end{itemize}

\begin{table}[t]
\centering
\tabfont
\caption{Engine changes, measured on one A100 (40\,GB) at the worker, excluding the network.}\label{tab:engine}
\begin{tabular*}{\linewidth}{@{\extracolsep{\fill}}llcc@{}}
\toprl
Change & Metric & Before & After \\
\midrule
Batched VAE decoding & VAE decode per step, 96 streams (ms) $\downarrow$ & 110 & \textbf{27} \\
 & Audio seconds per second, 48 streams $\uparrow$ & 41 & \textbf{74} \\
CUDA-graph prefill & First audio, 1 stream (ms) $\downarrow$ & 74 & \textbf{32} \\
 & First audio, 96 streams (ms) $\downarrow$ & 143 & \textbf{94} \\
Kernel warm-up & Stall on a new batch size (s) $\downarrow$ & $\sim$1.4 & \textbf{none} \\
\botrl
\end{tabular*}
\end{table}

\begin{figure}[t]
\begin{wide}
\centering
\ref{latleg}\\[1pt]
\begin{tikzpicture}
\begin{groupplot}[group style={group size=2 by 1,horizontal sep=1.7cm},paperaxis,scale only axis,width=6.3cm,height=3.9cm,
  xmode=log,log basis x=2,xtick={1,16,48,96},xticklabels={1,16,48,96},xlabel={Concurrent streams},xmajorgrids,ymajorgrids]
\nextgroupplot[ylabel={Median first audio (ms)},ymin=0,ymax=160,legend to name=latleg,legend columns=2,title={\textbf{a}\enspace Latency, lower is better},title style={at={(0,1)},anchor=south west,font=\sffamily\footnotesize,text=ink}]
\addplot[muted!70,mark=o,thick,mark options={fill=white}] coordinates {(1,74) (16,108) (48,125) (96,143)};
\addplot[cobalt,mark=*,thick] coordinates {(1,32) (16,62) (48,72) (96,94)};
\legend{Before CUDA-graph prefill,After}
\nextgroupplot[ylabel={Seconds of audio per second},ymin=0,ymax=150,title={\textbf{b}\enspace Throughput, higher is better},title style={at={(0,1)},anchor=south west,font=\sffamily\footnotesize,text=ink}]
\addplot[muted!70,mark=o,thick,mark options={fill=white}] coordinates {(1,10.2) (16,46) (48,74) (96,102)};
\addplot[cobalt,mark=*,thick] coordinates {(1,11.2) (16,69) (48,106) (96,136)};
\end{groupplot}
\end{tikzpicture}
\caption{Latency and throughput on one A100 as concurrent streams grow, before and after CUDA-graph prefill. Batched VAE decoding is on in both.}\label{fig:latency}
\end{wide}
\end{figure}
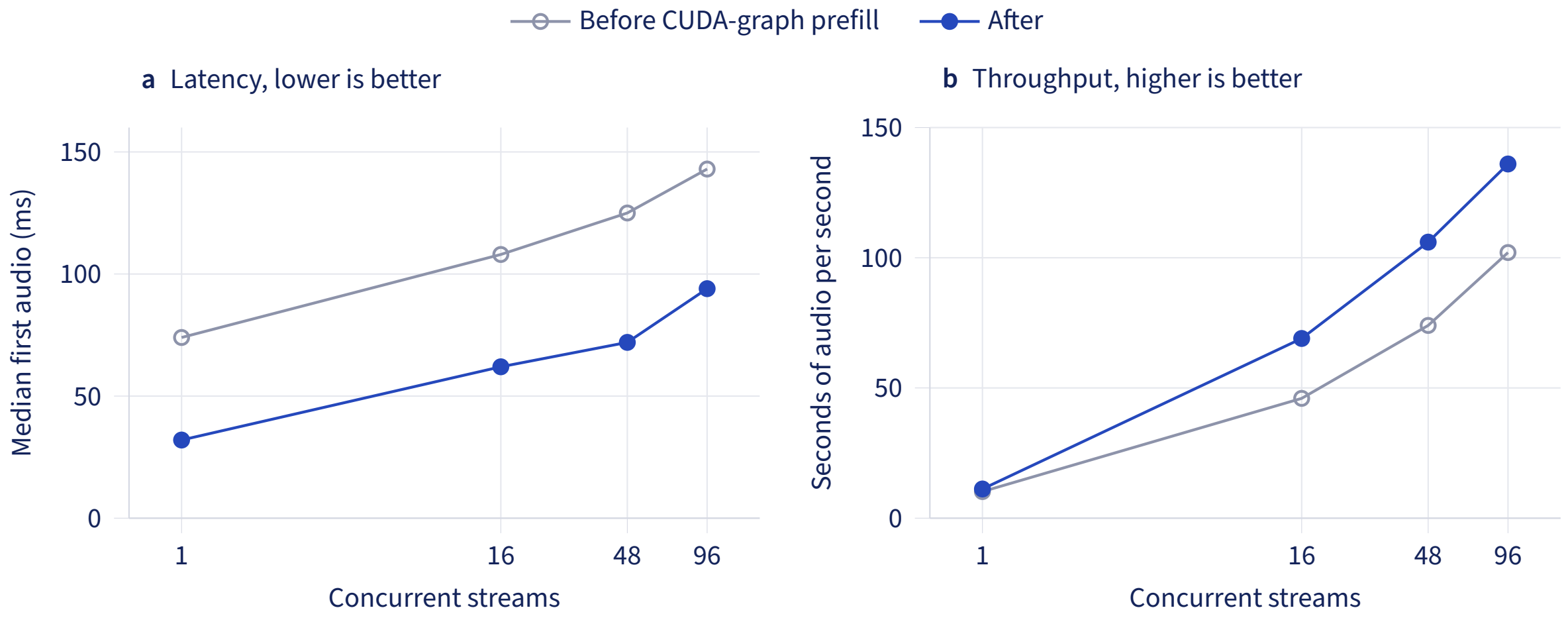

Figure~\ref{fig:latency} shows the effect of the prefill change as load grows, with batched decoding already in place. With all three changes, median first audio is 32\,ms for one stream and 94\,ms for 96 concurrent streams, which together produce 136 seconds of audio per second. Through the API on the same machine, a single stream starts in 97\,ms for a fixed voice, 109\,ms for a clone and 78\,ms for a designed voice. On an H200 with Poisson arrivals of $\sim$5-s requests, median first audio is 51\,ms at 2 requests per second and 75\,ms at 12 requests per second, where 98.6\% of streams still keep ahead of real time.

\subsection{Serving many fine-tuned voices}

Professional clones (Section~\ref{sec:cloning}) are LoRA adapters on one shared base model, as in S-LoRA \citep{sheng2023slora} and Punica \citep{chen2023punica}. Each request names its adapter. Up to eight adapters stay on the GPU, and requests with different adapters, or none, share a batch. Adapters load from storage on first use and are cached in host memory and on disk. A broken adapter falls back to the instant clone of the same voice.

\paragraph{Prefix-cache correctness.} The engine reuses cached key/value blocks when a new request starts with the same tokens as an earlier one. Its cache key covered only the tokens, so a request for a fine-tuned voice could reuse blocks that a plain request had computed from the same reference and text. We saw 8 of 9 prompt blocks reused this way, and 3 of 3 fine-tuned renders differed from renders with an empty cache. We now seed the cache key with the adapter's name and a digest of its weights. Fine-tuned renders then match empty-cache renders bit for bit, and plain requests keep their existing keys.

\paragraph{Two smaller fixes.} We replaced a LoRA kernel that would not compile on our Triton version; outputs agree to within $2\times10^{-7}$. Batches with a single adapter now run as plain matrix products, which cut the adapter's first-audio overhead from 33 to 13\,ms.

With these fixes, quality through the engine matches the reference implementation (Table~\ref{tab:multilora}). In mixed batches, every render was closest to its own speaker (60 of 60), and throughput with every stream on an adapter is 4--15\% below plain clones.

\begin{table}[t]
\centering
\tabfont
\caption{Professional clones through the engine (4 speakers $\times$ 5 texts $\times$ 3 seeds). Speaker similarity $\uparrow$ / CER $\downarrow$.}\label{tab:multilora}
\begin{tabular*}{\linewidth}{@{\extracolsep{\fill}}lcc@{}}
\toprl
Path & Instant clone & Professional clone \\
\midrule
Reference implementation & 0.651 / 0.019 & 0.770 / 0.026 \\
Our engine & 0.659 / 0.019 & 0.772 / 0.028 \\
Our engine, mixed batches & 0.676 / 0.023 & 0.780 / 0.026 \\
\botrl
\end{tabular*}
\end{table}

\section{Limitations}\label{sec:limitations}

Our evaluation has several gaps. First, naturalness rests on UTMOS22, a predictor trained on English speech, and on blind comparisons by one native listener. Second, intelligibility depends on the speech recognizer and on the text normalization used for scoring, and the same recognizers also helped build our training data. Third, ECAPA speaker similarity separates speakers on our data, but differences below about 0.05 should not be over-read: a style instruction alone moves similarity to the plain voice from about 0.91 to between 0.64 and 0.80. Fourth, most results come from a single seed or a single run.

The model itself has limits. Ninety percent of its training audio comes from VoxCPM2, so Alania-2 inherits its teacher's prosody and artefacts, including narrow high-frequency tones in some voices that we filter out of the synthetic data but do not yet suppress at serving time. The base training text was also normalized before the native-speaker review of Section~\ref{sec:normalization}, so the model now receives some readings, including all brand respellings, that it never saw in training. Finally, exact GPU-hours and some early training configurations were not recorded.

\section{Ethics and Safety}\label{sec:ethics}

\paragraph{Training data.}
The real speech comes from public YouTube videos distributed in YODAS under Creative Commons terms. Our synthetic text includes Wikipedia (CC BY-SA) and FineWeb2 (ODC-By). The built-in voices are designed voices that do not belong to any real person, and our released synthetic speech contains designed voices only. We do not release the model weights.

\paragraph{Cloning.}
Cloning someone's voice without their agreement can enable fraud, harassment and disinformation. Every clone requires the user to state that they have the speaker's consent, and we record this statement with the voice. Professional clones also require the live reading and speaker match of Section~\ref{sec:cloning}, with separate consent to the biometric comparison. Speaker embeddings are computed in memory and never stored, uploaded recordings are deleted seven days after training, customer voice data stays in the EU, and every response is labelled as AI-generated.

These controls have gaps. Instant clones have no live verification, so only the user's statement of consent stands in the way of misuse. We do not yet watermark generated audio \citep{sanroman2024audioseal}, and we have no specific policy on the voices of public figures beyond requiring consent. Every voice in this report's cloning experiments belongs to a consenting contributor or comes from FLEURS.

\section{Conclusion}\label{sec:conclusion}

We presented Alania-2, a 488M-parameter Turkish TTS model trained mostly on synthetic speech. On its built-in voices, it comes within half a point of CER of a fine-tuned model more than four times its size, generates 48\,kHz audio, and starts streaming in tens of milliseconds. Larger models bought little intelligibility, and a 7.6M-parameter model trained with the same recipe matched the best systems we measured; extra parameters pay for flexibility rather than accuracy. Several of the most useful lessons concerned the system around the model. Alania-2, trained on normalized text, needs a text normalizer to read digits and abbreviations. Cloning by continuing a reference fails with long references, while conditioning on the reference audio alone does not, and 20--30 seconds of reference audio works best. A low-rank adapter trained in about two minutes gives a close voice match, and serving many such adapters needs care with caching. Next, we plan a listening test with many native listeners, audio watermarking, live verification for instant clones, and retraining on text from the current normalizer.

\section*{Data and Code Availability}
\addcontentsline{toc}{section}{Data and Code Availability}
The Freya-TR-Eval harness and per-system results are public at \url{https://huggingface.co/datasets/PatientdeskAI/freya-tr-eval-rerun}. The caption, synthetic speech and domain text datasets are linked in Section~\ref{sec:data}. Antalia-2 Mini is public at \url{https://huggingface.co/cloud0day3/antalia-mini}. The weights of Alania-2 and Duyu and our serving code are not public; both models can be used through our speech API at \url{https://speech.patientdesk.ai}.

\section*{Competing Interests}
\addcontentsline{toc}{section}{Competing Interests}
The authors are affiliated with PatientDesk AI, which offers Alania-2 and Duyu through a commercial speech API and develops Antalia-2 Mini. Duyu, used for filtering and in some evaluations in this report, is the authors' own model.

\bibliographystyle{plainnat}
\bibliography{refs}

\keepspace{0.6\textheight}
\appendix
\clearpage
\appendix
\section{Full Candidate Results}\label{app:tracks}

\begin{table}[!h]
\begin{wide}
\centering
\tabfont
\caption{Candidates on the 292-clip suite (one seed). CER from Whisper large-v3, UTMOS22, ECAPA similarity to the voice reference; ``gender'' counts voice designs that matched the requested gender.}
\begin{tabular*}{\linewidth}{@{\extracolsep{\fill}}lcccccccccc@{}}
\toprl
& \multicolumn{3}{c}{Female voice} & \multicolumn{3}{c}{Male voice} & \multicolumn{3}{c}{Style} & Design \\
\cmidrule(lr){2-4}\cmidrule(lr){5-7}\cmidrule(lr){8-10}\cmidrule(lr){11-11}
Model & CER & UTMOS & Sim & CER & UTMOS & Sim & CER & UTMOS & Sim & Gender \\
\midrule
VoxCPM2, zero-shot (2.3B) & .027 & 3.77 & .748 & .029 & 3.53 & .812 & .022 & 2.96 & .542 & 23/32 \\
Track 1, round 1 (2.3B) & .028 & 3.77 & .690 & .027 & 3.54 & .806 & .022 & 3.66 & .667 & 25/32 \\
Track 1, round 1 LoRA & .024 & 3.78 & .707 & .028 & 3.60 & .810 & .037 & 3.71 & .714 & 23/32 \\
Track 1, round 2 (2.3B) & .024 & 3.73 & .761 & .023 & 3.50 & .737 & .020 & 3.13 & .534 & 27/32 \\
Track 2, first run (488M) & .061 & 3.48 & .768 & .055 & 3.12 & .723 & .142 & 3.21 & .663 & 30/32 \\
Track 2, base 70k (488M) & .027 & 3.64 & .773 & .028 & 3.28 & .739 & .090 & 3.29 & .666 & 24/32 \\
\textbf{Alania-2} (fine-tune 4k) & .027 & 3.57 & .766 & .028 & 3.24 & .741 & .034 & 3.23 & .633 & 30/32 \\
Fine-tune 12k & .027 & 3.62 & .769 & .028 & 3.26 & .736 & .044 & 3.22 & .612 & 30/32 \\
\botrl
\end{tabular*}
\end{wide}
\end{table}

\section{Systems in the Freya-TR-Eval Re-run}\label{app:systems}

\begin{table}[!h]
\begin{wide}
\centering
\tabfont
\caption{Systems in our Freya-TR-Eval re-run (Figure~\ref{fig:freya}), with the exact checkpoint or service we ran. Model identifiers are Hugging Face repositories unless noted. Parameter counts are those in our released results; ``--'' means not published.}\label{tab:systems}
\begin{tabular*}{\linewidth}{@{\extracolsep{\fill}}l r >{\raggedright\arraybackslash}p{6.0cm} l@{}}
\toprl
System & Params & Checkpoint or service & Paper \\
\midrule
EMA Lightning & 8.6M & \texttt{ema-lightning} 1.0.1 (PyPI), \href{https://huggingface.co/canberkkkkkk/ema-lightning}{\texttt{canberkkkkkk/ema-lightning}} & -- \\
Antalia-2 Mini & 7.62M & 1.0.0, \href{https://huggingface.co/cloud0day3/antalia-mini}{\texttt{cloud0day3/antalia-mini}} & Section~\ref{sec:size} \\
Trendyol-TTS & 2.38B & \href{https://huggingface.co/Trendyol/Trendyol-TTS}{\texttt{Trendyol/Trendyol-TTS}} & -- \\
Gemini 3.8 Flash TTS & -- & Vertex AI, \texttt{gemini-3.8-flash-tts}, voice Kore & -- \\
Gemini 3.8 Flash-Lite TTS & -- & Vertex AI, \texttt{gemini-3.8-flash-lite-tts}, voice Kore & -- \\
Anka TTS & 336M & \texttt{anka-tts} v0.1 (PyPI) & -- \\
Alania-2 & 488M & our API, and the bare model & this report \\
Chatterbox Multilingual & 500M & \href{https://github.com/resemble-ai/chatterbox}{\texttt{resemble-ai/chatterbox}} (GitHub) & -- \\
Piper & 16M & \texttt{tr\_TR-dfki-medium} from \href{https://huggingface.co/rhasspy/piper-voices}{\texttt{rhasspy/piper-voices}} & -- \\
XTTS-v2 & 470M & \href{https://huggingface.co/coqui/XTTS-v2}{\texttt{coqui/XTTS-v2}} & \citet{casanova2024xtts} \\
Alania-1 & 3.3B & our API & -- \\
Antalia-1 & 305M & \href{https://huggingface.co/cloud0day3/antalia-1}{\texttt{cloud0day3/antalia-1}} & -- \\
MMS-TTS & 36M & \href{https://huggingface.co/facebook/mms-tts-tur}{\texttt{facebook/mms-tts-tur}} & \citet{pratap2023mms} \\
Coqui GlowTTS & 28M & \texttt{tts\_models/tr/common-voice/} \texttt{glow-tts} (Coqui TTS) & \citet{kim2020glowtts} \\
FreyaTTS-small & 183M & \href{https://huggingface.co/freyavoice/Freya-TTS}{\texttt{freyavoice/Freya-TTS}} & \citet{pamuk2026freyatts} \\
\botrl
\end{tabular*}
\end{wide}
\end{table}

\end{document}